\documentclass{article}

\usepackage{fullpage, physics, amsmath, amssymb, amsthm, algorithm, algorithmic, graphicx, url, xcolor, comment, appendix, caption, subcaption, braket, bm, here, url, tabularx, booktabs, array, float, subcaption}

\title{A Quantum-Inspired Algorithm for Solving Sudoku Puzzles and the MaxCut Problem\thanks{Received on June 5, 2025, accepted on August 6, 2025, Quantum Information \& Computation\\ISSN 1533-7146 Volume 25 (2025) pp.xx--xx}}

\author{Max B. Zhao\thanks{Thomas Jefferson High School for Science and Technology, Alexandria, VA 22312, USA. Email: 2026mzhao@tjhsst.edu. This work was initiated during Max B. Zhao's 2024 ASSIP internship at George Mason University.} \and Fei Li\thanks{Department of Computer Science, George Mason University, Fairfax, VA 22030; Email: fli4@gmu.edu}}

\date{June 2025}

\begin{document}

\maketitle


\begin{abstract}
We propose and evaluate a quantum-inspired algorithm for solving Quadratic Unconstrained Binary Optimization (QUBO) problems, which are mathematically equivalent to finding ground states of Ising spin-glass Hamiltonians. The algorithm employs Matrix Product States (MPS) to compactly represent large superpositions of spin configurations and utilizes a discrete driving schedule to guide the MPS toward the ground state. At each step, a driver Hamiltonian -- incorporating a transverse magnetic field -- is combined with the problem Hamiltonian to enable spin flips and facilitate quantum tunneling. The MPS is updated using the standard Density Matrix Renormalization Group (DMRG) method, which iteratively minimizes the system's energy via multiple sweeps across the spin chain. Despite its heuristic nature, the algorithm reliably identifies global minima, not merely near-optimal solutions, across diverse QUBO instances. We first demonstrate its effectiveness on intermediate-level Sudoku puzzles from publicly available sources, involving over $200$ Ising spins with long-range couplings dictated by constraint satisfaction. We then apply the algorithm to MaxCut problems from the Biq Mac library, successfully solving instances with up to $251$ nodes and $3,265$ edges. We discuss the advantages of this quantum-inspired approach, including its scalability, generalizability, and suitability for industrial-scale QUBO applications.
\end{abstract}



\section{Introduction}
\label{sec:intro}


Combinatorial optimization problems are both fundamentally important and computationally challenging
~\cite{papadimitriou1998combinatorial, korte2011combinatorial}. They arise in a wide range of industries, including logistics, supply chain management, vehicle routing, drug discovery, and portfolio optimization. Classical problems such as the traveling salesman problem have played a pivotal role in shaping the field of computational complexity theory~\cite{cormen2022introduction}. A broad class of these optimization problems can be reformulated as Quadratic Unconstrained Binary Optimization (QUBO) problems~\cite{Glover:2022aa, punnen2022quadratic}, which admit a compact mathematical representation:
\begin{align*}
\min & f = \sum_{i, j} x_i Q_{i, j} x_j & x_i \in \{0, 1\}
\end{align*}
where $Q$ is a symmetric matrix. Despite its seemingly simple form, the QUBO problem is NP-hard~\cite{Barahona_1982}, meaning that no classical algorithm can efficiently solve all instances as the problem size increases unless P = NP~\cite{cormen2022introduction}. Consequently, many industrial-scale QUBO problems pose significant computational challenges~\cite{punnen2022quadratic}. While classical heuristic algorithms remain the primary tools for obtaining near-optimal solutions, there is growing interest in quantum and quantum-inspired approaches. These emerging methods hold the promise of improved scalability and solution quality, particularly for large and complex problem instances.

A common strategy in quantum optimization algorithms is to encode the objective function into a cost Hamiltonian $H_c$, which is then manipulated using quantum superposition, tunneling, and other non-classical effects~\cite{Abbas:2024aa}. Since the variables $x_i$ are binary, a QUBO problem can be naturally mapped onto an Ising-type spin (or qubit) Hamiltonian~\cite{10.3389/fphy.2014.00005, Mohseni:2022aa}. In this mapping, the QUBO matrix $Q$ defines the Ising couplings, and the cost function $f$ corresponds to the eigenvalues of $H_c$. Thus, solving the optimization problem becomes equivalent to finding the ground state of $H_c$. In quantum annealing~\cite{finnila1994quantum, PhysRevE.58.5355, RevModPhys.80.1061}, a driver Hamiltonian that does not commute with $H_c$ is introduced. The system is initialized in the ground state of the driver Hamiltonian---typically an equal-weight superposition over all computational basis states---and is then evolved adiabatically as the driver strength is gradually decreased and the cost Hamiltonian is amplified. Under ideal conditions, the system remains in its instantaneous ground state throughout this process, eventually reaching the ground state of $H_c$, which encodes the optimal solution. The Quantum Approximate Optimization Algorithm (QAOA)~\cite{farhi2014quantum, hadfield2019quantum} takes a different approach by applying alternating layers of unitary transformations -- referred to as \emph{cost and mixer layers} -- to a set of qubits initialized in some superposition state. After each iteration, the expectation value of the cost Hamiltonian is measured, and a classical optimizer updates the variational parameters to maximize performance. This hybrid quantum-classical loop is repeated until a set of parameters is found that yields a high-quality solution~\cite{hadfield2019quantum}. In our quantum-inspired algorithm, we adopt the concepts of driver and mixer Hamiltonians to guide the optimization process in a purely classical framework.

The practical implementation of quantum algorithms is currently constrained by the limitations of available hardware. For example, in the D-Wave Advantage annealer, each qubit can couple with at most 15 other qubits~\cite{advantage}, whereas many QUBO problems require denser or long-range connectivity. Similarly, the applicability of QAOA is limited by qubit coherence times and circuit depth, restricting its use to relatively small systems with up to approximately $40$ qubits~\cite{pagano2020quantum, blekos2024review}. These hardware constraints have spurred interest in quantum-inspired algorithms, which do not rely on quantum hardware but instead simulate certain quantum behaviors---such as superposition and tunneling -- on classical machines. By emulating these features, quantum-inspired approaches have demonstrated the potential to outperform traditional classical heuristics on a variety of optimization problems~\cite{portfolio, pittmine, inspiredreview, du2025new}.

In this work, we propose and evaluate a quantum-inspired algorithm for solving QUBO problems involving over $200$ variables with dense connectivity. To benchmark its performance, we focus on problem instances where the optimal solution is challenging to find but straightforward to verify. Although many curated optimization problems exist, the global optimum is often unknown or unproven for medium- to large-scale instances. This limitation motivates our choice of a well-established class of problems that admit QUBO formulations, offer multiple variants, and critically have easily verifiable solutions: Sudoku puzzles.

This work is motivated by a simple question: Can quantum mechanics help solve Sudoku puzzles? Although Sudoku can be formulated as a QUBO problem, it appears to be beyond the practical reach of current quantum optimization methods. For instance, M\"{u}cke~\cite{mucke} reports that quantum annealing was unable to reliably find the correct solution (see Figure 4 in~\cite{mucke}), and QAOA has not demonstrated success on such instances. In this sense, Sudoku puzzles represent a kind of ``worst-case scenario'' for QUBO problems: the cost function is entirely constraint-driven, with non-local dependencies involving many binary variables. Once the puzzle's clues are incorporated, the resulting Ising model exhibits long-range couplings and an inhomogeneous on-site magnetic field -- features that make the spin landscape highly complex and challenging to optimize. Moreover, Sudoku provides an exacting benchmark: the solution is easy to verify, but correctness demands that the algorithm locate the exact ground state. Merely reducing the energy or producing a partially correct board is insufficient. As such, success requires not just approximate optimization but full satisfaction of all constraints. This makes Sudoku puzzles an ideal testbed for assessing the reliability and quality of a QUBO solver.


\subsection{Our Contribution}

We emphasize that the goal of this work is not to develop a Sudoku solver that outperforms classical methods. Indeed, Sudoku can be solved manually, or via efficient classical algorithms such as backtracking, constraint propagation, or simulated annealing applied to its QUBO formulation. Rather, our interest in Sudoku lies in its ability to generate hard, structured QUBO problems that have proven challenging for quantum and quantum-inspired approaches. Once our algorithm successfully passes this stringent test, we extend its application to a range of MaxCut problems. Although MaxCut instances differ in structure from Sudoku, the algorithm continues to perform well, consistently finding optimal solutions.

A central innovation of our approach is the combination of a discrete driving schedule with a sweeping procedure applied to the many-spin wavefunction. Specifically, we adapt techniques from quantum many-body physics, borrowing the concepts of Matrix Product States (MPS)~\cite{verstraete} and the Density Matrix Renormalization Group (DMRG) algorithm~\cite{white, dmrgreview}, both of which originate in the study of low-dimensional quantum systems~\cite{tensorreview}. While MPS and DMRG were originally developed for one-dimensional systems with local interactions, our proposed ``hop, sweep, repeat'' procedure is capable of efficiently converging to the ground state of spin systems derived from QUBO instances with inhomogeneity, frustration, and long-range couplings. The insights gained from these case studies form a foundation for applying our method to broader classes of combinatorial optimization problems.


\subsection{Paper Organization} 

This paper is structured as follows. Section~\ref{sec:sudoku} introduces the formulation of the Sudoku problem as an Ising spin system and analyzes the structure of the resulting spin model. Section~\ref{sec:algs} presents our quantum-inspired algorithm in detail. The algorithm’s performance is evaluated in Section~\ref{sec:experimentS} and Section~\ref{sec:experimentM} for Sudoku and MaxCut problems, respectively. Additional discussion and concluding remarks are provided in Section~\ref{sec:conclusions}. Further technical details about MPS, DMRG, and the search strategy are provided in Appendices~\ref{appendix:A} and~\ref{appendix:B}.


\section{Formulating Sudoku as Ising Spin Glass}
\label{sec:sudoku}

Sudoku is a number puzzle game enjoyed by players around the world. Daily Sudoku challenges are featured in major newspapers, and puzzle books are commonly found at airport newsstands. The standard Sudoku board consists of a $9 \times 9$ grid, divided into nine $3 \times 3$ blocks. The objective is to fill the grid with digits from $1$ to $9$ such that each row, column, and block contains every digit exactly once. The puzzle begins with a set of prefilled cells, known as \emph{clues}. Generally, puzzles with fewer clues are more difficult to solve. Although Sudoku can be challenging, it can be solved efficiently using classical algorithms, such as backtracking.

The Sudoku grid can be generalized to an $n^2 \times n^2$ structure, where $n$ is the size of each block ($n = 3$ corresponds to the standard Sudoku puzzle). This generalization positions Sudoku as a compelling example in the study of computational complexity. Solving generalized Sudoku has been proven to be NP-hard~\cite{yato2003complexity}, implying that the problem becomes computationally intractable for classical algorithms as $n$ increases. Moreover, Ueda and Nagao~\cite{UedaN96} showed that Sudoku is ASP-complete (Another Solution Problem complete), meaning that even deciding whether a given instance has more than one solution is computationally hard. These results firmly establish Sudoku within the class of difficult combinatorial problems and motivate its investigation through alternative, including quantum-inspired, computational approaches.

To place Sudoku within the broader class of well-known NP-hard problems, it is useful to cast it as an optimization task. In this section, we introduce a straightforward formulation of Sudoku as a QUBO problem. While this approach is not optimized for computational efficiency and results in a large number of binary variables, it effectively reveals the structural parallels between Sudoku and other combinatorial optimization problems. As a result, quantum-inspired algorithms designed for solving Sudoku can be readily adapted to tackle a wide range of related optimization challenges.


\subsection{QUBO Formulation}


Our QUBO formulation closely follows the approach introduced by Mücke~\cite{mucke}, though our implementation of the constraints and cost function differs in key ways. Let $z_{i, j, k}$ be a binary variable that takes the value $1$ if the cell in the $i$th row and $j$th column of the Sudoku grid is assigned the number $k$, and $0$ otherwise, where $i, j, k = 1, 2, \ldots, 9$. The rules of Sudoku can then be encoded through the following four sets of constraints:
\begin{eqnarray}
\sum_k z_{i, j, k} = 1, & \forall i, j \label{eq1}\\
\sum_i z_{i, j, k} = 1, & \forall j, k \label{eq2}\\
\sum_j z_{i, j, k} = 1, & \forall i, k \label{eq3}\\
\sum_{(i, j) \in b} z_{i, j, k} = 1, & \forall b, k \label{eq4}
\end{eqnarray}  

The first constraint~\eqref{eq1} enforces that each cell $(i, j)$ must be assigned exactly one number. The second constraint~\eqref{eq2} ensures that, within each column $j$, a given number $k$ appears only once. Similarly, constraint~\eqref{eq3} requires that each number $k$ appears exactly once in every row $i$. The fourth constraint~\eqref{eq4} sums over all cells $(i, j)$ within the same block $b$ to enforce that each number $k$ occurs only once per block.

Following standard practice~\cite{Glover:2022aa}, we convert these constraints into penalty terms, which are incorporated into the overall cost function. For instance, the penalty corresponding to constraint~\eqref{eq1} is given by:
\begin{align}
\nonumber P_1 & = \lambda \sum_{i, j} \left(\sum_k z_{i, j, k} - 1\right)^2\\
\nonumber & = \lambda \sum_{i, j} \left(1 - 2 \sum_k z_{i, j, k} + \left(\sum_k z_{i, j, k}\right)^2\right)\\
\label{eq:P0} & = \lambda \sum_{i, j} \left(1 - 2 \sum_k z^2_{i, j, k} + \sum_k z_{i, j, k}^2 + \sum_k \sum_{k' \neq k} z_{i, j, k} z_{i, j, k'}\right)\\
\label{P1} & = \lambda \sum_{i, j} \left(1 - \sum_k z^2_{i, j, k} + \sum_k \sum_{k' \neq k} z_{i, j, k} z_{i, j, k'}\right)
\end{align}  

Note that Equation~(\ref{eq:P0}) holds due to the property $x = x^2$ for a binary variable $x$. We observe that the terms within the parentheses in~\eqref{P1} are either constants or quadratic in $z$, and the off-diagonal coupling is dense -- each $z_{i, j, k}$ is coupled to every $z_{i, j, k' \neq k}$. The other penalties, $P_2$ through $P_4$, can be applied in a similar manner for constraints~\eqref{eq2} through~\eqref{eq4}. The cost function we seek to minimize is simply the sum:
\begin{equation}
f(\{z_{i, j, k}\}) = P_1 + P_2 + P_3 + P_4
\end{equation}

For simplicity, we assign the same penalty weight $\lambda > 0$ to all four constraint terms. The binary variables $z_{i, j, k}$ can then be organized into a single vector $Z$, allowing the total cost function to be expressed compactly in quadratic form. 

\begin{equation}
Z_{81 i + 9 j + k} = z_{i,j,k}
\end{equation}

Accordingly, the cost function $f$ assumes the standard form
\begin{equation}
f\left(\{Z_m\}\right) = \lambda \left[  c_1+  \sum_{m, n = 1}^{9^3} Z_m Q_{m, n} Z_n\right]
\label{fatqubo}
\end{equation}
where the constant $c_1 = 4 \times 9^2$, and the QUBO matrix $Q$ is constructed by collecting both the diagonal and off-diagonal couplings -- such as those shown in~\eqref{P1} -- among the elements of $Z$. The QUBO formulation introduces a large number of binary variables (to be reduced in the next step), organized into the vector $Z$. Unlike many typical QUBO problems, the cost function in this case arises solely from constraints. Clearly, the global minimum of $f$ is $0$, as any valid solution satisfies all constraints, causing all four penalty terms $P$s to vanish. 

Due to our uniform choice of penalty structure, the cost function $f$ is proportional to $\lambda$. From this point forward, we measure $f$ in units of $\lambda$, effectively setting $\lambda = 1$ so that it drops out of the formulation. The clues provided by each Sudoku puzzle fix the values of certain variables $z_{i, j, k}$, significantly reducing the number of free variables from the original $9^3$. This reduction process is known as \emph{clamping}, as described by Mücke~\cite{mucke}. Below, we briefly summarize the procedure using our notation. Let $N_c$ denote the number of clues, i.e., the number of pre-filled cells. For each cell $(i^*, j^*)$ with a given value $k^*$, we first set $z_{i^*, j^*, k^*} = 1$, and then propagate this clue by applying the constraints~\eqref{eq1} through~\eqref{eq4}, setting
\begin{align}
z_{i^*, j^*, k \neq k^*} = 0\\
z_{i \neq i^*, j^*, k^*} = 0\\
z_{i^*, j \neq j^*, k^*} = 0\\
z_{(i, j) \in b^*, k^*} = 0
\end{align}
where $(i, j) \in b^*$ denotes all other cells within the same block that contains $(i^*, j^*)$. After all clues have been applied and the affected $z$-values are ``clamped'' to either $1$ or $0$, we collect the remaining free binary variables and organize them into a new vector $X$. The size of $X$, denoted $N_s$, is smaller than that of the original vector $Z$. However, there is no simple relationship between $N_c$ (the number of clues) and $N_s$. During the propagation process, some variables $z_{i, j, k}$ may be clamped to zero multiple times, so $N_s$ depends not only on the number of clues but also on their specific positions and values. This clamping procedure effectively reduces the size of the QUBO problem
\begin{align*}
Z \rightarrow X, & & Q \rightarrow J    
\end{align*}
and the QUBO problem~\eqref{fatqubo} reduces to 
\begin{equation}
f\left(\{X_m\}\right) =  c_1 + c_2 + \sum^{N_s}_{m, n = 1} X_m J_{m, n} X_n
\label{leanqubo}
\end{equation}

The second constant $c_2$ emerges as a result of the clamping process, and the reduced QUBO matrix $J$ is derived from the original matrix $Q$ via an appropriate transformation.

\begin{displaymath}
J = T^\top Q T    
\end{displaymath}

More details about the transformation matrix $T$ and the constant $c_2$ can be found in~\cite{mucke}. For an intermediate-level Sudoku puzzle with $24$ to $26$ clues, the reduced QUBO problem typically has $N_s$ exceeding $200$. While this is much smaller than the original $9^3$ variables, it remains significantly larger than the problem sizes commonly addressed in typical QAOA applications reported in the literature.

Every QUBO problem can be mapped to an Ising-type classical spin model. For each binary variable $X_m$, we define a corresponding Ising spin-$z$ variable
\begin{equation}
S^z_m = X_m - \frac{1}{2}
\label{x-to-s}
\end{equation}

It takes the value $+\frac{1}{2}$ (referred to as spin-up, $\uparrow$) when $X_m = 1$, and $-\frac{1}{2}$ (spin-down, $\downarrow$) when $X_m = 0$. After this change of variables, the cost function $f$ is reinterpreted as the Hamiltonian (i.e., the energy function) of the spin model. Note $\left(S^z_m\right)^2 = \frac{1}{4}$.

\begin{align}
\nonumber H_z  & := f\left(\{X_m\}\right)\\
\nonumber & = c_1 + c_2 + \sum^{N_s}_{m, n = 1} X_m J_{m, n} X_n\\
\nonumber & = c_1 + c_2 + \sum^{N_s}_{m, n = 1} \left(S^z_m + \frac{1}{2}\right) J_{m, n} \left(S^z_n + \frac{1}{2}\right)\\
\nonumber & = c_1 + c_2 + \sum^{N_s}_{m, n = 1} S^z_m J_{m, n} S^z_n + \frac{1}{2} \sum^{N_s}_{m, n = 1} S^z_m J_{m, n} + \frac{1}{2} \sum^{N_s}_{m, n = 1} J_{m, n} S^z_n + \frac{1}{4} \sum^{N_s}_{m, n = 1} J_{m, n}\\
\nonumber & = c_1 + c_2 + \left(\sum^{N_s}_{m = 1} J_{m, m} (S^z_m)^2 + \sum^{N_s}_{m \neq n} S^z_m J_{m, n} S^z_n\right) + \frac{1}{2} \sum^{N_s}_{m, n = 1} \left(S^z_m J_{m, n} + J_{m, n} S^z_n\right) + \frac{1}{4} \sum^{N_s}_{m, n = 1} J_{m, n}\\
\nonumber & = c_1 + c_2 + \left(\frac{1}{4} \sum^{N_s}_{m = 1} J_{m, m} + \sum^{N_s}_{m \neq n} S^z_m J_{m, n} S^z_n\right) + \frac{1}{2} \sum^{N_s}_{m = 1} \sum^{N_s}_{n = 1} \left(S^z_m J_{m, n} + J_{m, n} S^z_n\right) + \frac{1}{4} \sum^{N_s}_{m, n = 1} J_{m, n}\\
& = c_1 + c_2 + c_3 + \sum^{N_s}_{m \neq n} S^z_m J_{m, n} S^z_n + \sum^{N_s}_{m = 1} h^z_m S^z_m
\label{SGH}
\end{align}
where the constant term $c_3$ is defined by 
\begin{equation}
c_3 =  \frac{1}{4} \sum^{N_s}_{m = 1} J_{m, m} +\frac{1}{4} \sum^{N_s}_{m, n=1} J_{m, n} 
\end{equation}
This shift can be traced back to the offset of $-\frac{1}{2}$ in equation~\eqref{x-to-s} as well as the self-interaction terms $J_{m, m} \left(S^z_m\right)^2$ in the summation with $\left(S^z_m\right)^2 = \frac{1}{4}$. In addition to the Ising interaction $J_{m, n}$ that couple spin $m$ and spin $n$, the Hamiltonian $H_z$ also includes an inhomogeneous (i.e., $m$-dependent) magnetic field in the $z$-direction that couples linearly to $S^z_m$.
\begin{equation}
h^z_m = \frac{1}{2} \sum^{N_s}_{n = 1} \left(J_{m, n} + J_{n, m}\right)
\end{equation}

The Hamiltonian $H_z$ contains $N_s$ spins in total. Although the spin index $m$ originates from the variable $z_{i, j, k}$, which indicates whether cell $(i, j)$ contains the number 
$k$, the spins themselves do not reside on the Sudoku board. Instead, it is useful to imagine the spins arranged along a fictitious one-dimensional chain, with sites indexed by $m$. At each site $m$, the local magnetic field takes the value $h^z_m$, and spins at sites $m$ and $n$ interact with strength $J_{m, n}$. This spin-chain picture is a convenient abstraction for describing and analyzing the problem. However, the chain geometry itself is not physically meaningful -- any graph that preserves the correct vertex weights ($h^z_m$) and edge weights ($J_{m, n}$) constitutes a valid representation of the spin system described by Hamiltonian~\eqref{SGH}. Following convention, we refer to Hamiltonian~\eqref{SGH} loosely as an Ising spin glass. Strictly speaking, however, the term ``spin glass'' applies to models where the couplings $J_{m, n}$ are randomly drawn from specific statistical distributions. In contrast, the $J$ matrix derived from Sudoku is fully determined by the game's rules and clues, and is therefore non-random.

Solving the Sudoku puzzle is now equivalent to finding the ground state of the Hamiltonian~\eqref{SGH} -- that is, identifying the spin configuration that minimizes the energy of the $N_s$ interacting spins in a magnetic field. A correct solution corresponds to zero energy and features exactly $(81 - N_c)$ spins pointing up. While the spin formulation offers a new perspective, it does not simplify the problem: finding the ground state of an Ising spin glass remains NP-hard. However, this formulation opens the door to developing quantum and quantum-inspired algorithms. To explore this direction, we generalize $H_z$ by adding a transverse magnetic field $h^x$ in the $x$-direction that couples to the spin-$x$ operator $S^x$, leading to the total Hamiltonian:
\begin{equation}
H_{\mbox{total}} = a H_x + b H_z
\label{H_total}
\end{equation}

Here, $a$ and $b$ are real control parameters to be specified. We allow the transverse field $h^x$ to be site dependent,
\begin{equation}
\nonumber
H_x = \sum_m h^x_m S^x_m 
\end{equation}
The spin operators are defined in the standard way. In the eigenbasis of $S^z$, they take the matrix form
\begin{equation}
S^z = \frac{1}{2}
\begin{pmatrix}
1 & 0\\
0 & -1
\end{pmatrix}
,\quad 
S^x = \frac{1}{2}
\begin{pmatrix}
0 & 1\\
1 & 0
\end{pmatrix}
\end{equation}
To simply the notation, we set the reduced Planck constant $\hbar = 1$ throughout.

To summarize, we have promoted the classical spin model~\eqref{SGH} to a quantum spin model~\eqref{H_total}. We refer to the term $H_x$ as the driver Hamiltonian because the transverse field drives the spins to precess and flip. In other words, it induces quantum tunneling between the spin-up and spin-down states, $\ket{\uparrow} \leftrightarrow \ket{\downarrow}$. Consequently, the spin state can exist in superpositions such as
\begin{displaymath}
\ket{\pm} = \frac{1}{\sqrt{2}} \left(\ket{\uparrow} \pm \ket{\downarrow}\right)
\end{displaymath}
and two spins can become entangled through the interplay of the Ising coupling and the transverse field $h^x$. To understand how the driver Hamiltonian $H_x$ can be leveraged to solve the spin problem, it is essential to characterize the properties of the coupling matrix $J_{m, n}$ and the local fields $h^z_m$.


\subsection{Characterization of the Ising Spin Glass}


It is instructive to compare the spin Hamiltonian~\eqref{SGH}, derived from Sudoku, with a well-known model of spin glass -- the Sherrington-Kirkpatrick Hamiltonian~\cite{PhysRevLett.35.1792}:
\begin{displaymath}
H_{K S} = h^z \sum_m S^z_m + \sum_{m \neq n} J_{m, n} S^z_m S^z_n     
\end{displaymath}

In $H_{K S}$, the magnetic field $h^z$is homogeneous across all sites, and the couplings $J_{m, n}$ are defined for every pair of spins $m$ and $n$ (i.e., ``all-to-all'' coupling), with values randomly drawn from a Gaussian distribution.

\begin{figure}[H]
\centering
\includegraphics[width=0.7\textwidth]{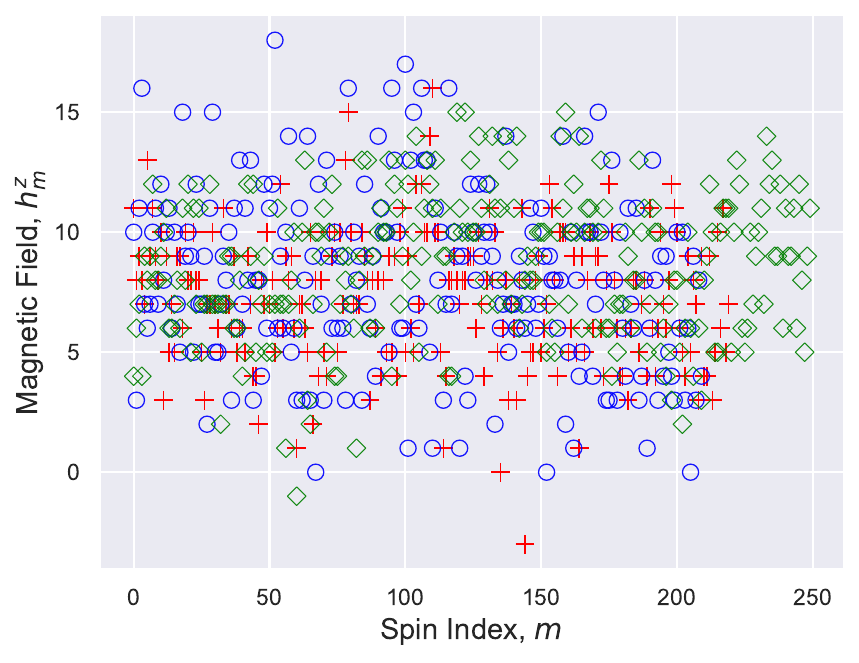}
\caption{The spin representations of three Sudoku puzzles published in The New York Times on January 2 (diamond), January 12 (circle), and January 14 (plus), 2025, are shown. The mean and standard deviation of the magnetic field $h^z_m$ are as follows: $8.7 \pm 3.0$ for the January 2 puzzle (diamond), $8.0 \pm 3.7$ for January 12 (circle), and $7.3 \pm 2.7$ for January 14 (plus).}
\label{fig:hzm}
\end{figure}

In contrast, for Sudoku-derived spin systems, the longitudinal magnetic field $h^z$ fluctuates significantly from site to site. Figure~\ref{fig:hzm} shows the site-dependent values $h^z_m$ for three intermediate-level Sudoku puzzles published in The New York Times on January 2 (diamonds), January 12 (circles), and January 14 (pluses), 2025. In all cases, the field exhibits highly irregular, large-amplitude variations over a broad range. According to the term $\sum_m h^z_m S^z_m$ in $H_z$, spins at sites with large positive $h^z_m$ favor pointing downward (i.e., $S^z_m = -\frac{1}{2}$), while those at sites with smaller or negative $h^z_m$ prefer to point upward. This ``rugged'' landscape of $h^z$, as visualized in Figure~\ref{fig:hzm}, leads to spin configurations that may appear random—markedly different from the ordered ferromagnetic or antiferromagnetic patterns seen in simpler spin models.

\begin{figure}[H]
\centering
\includegraphics[width=0.65\textwidth]{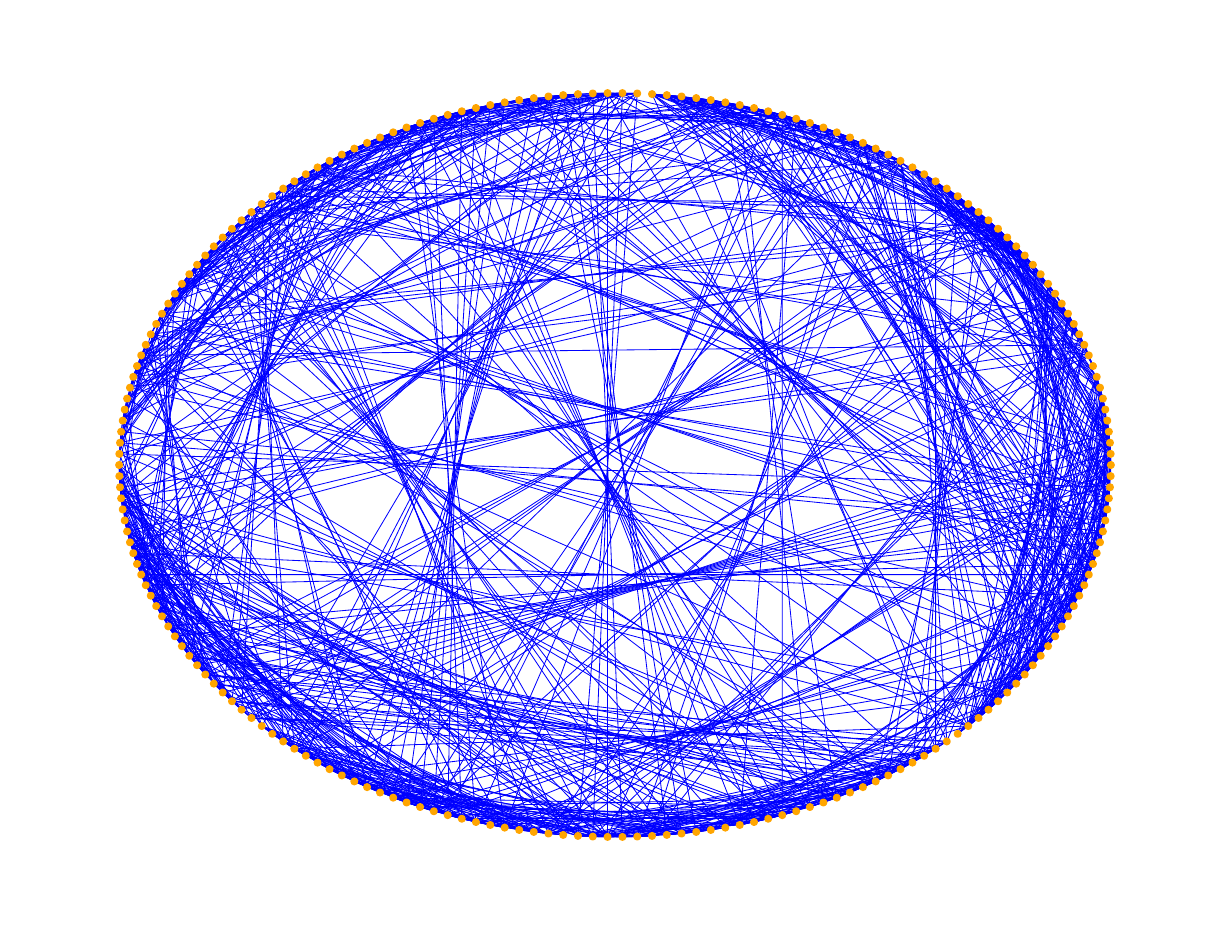}
\caption{The dense, long-range couplings among the 210 spins for the January 12, 2025 Sudoku puzzle. Each line represents an Ising coupling $J_{m, n} = 2$, which favors spin-$m$ and and spin-$n$ being antiparallel. Each spin can point either up or down. Finding the lowest-energy spin configuration corresponds to solving the Sudoku puzzle.}
\label{fig:conn}
\end{figure}

However, the minimization problem involves more than just the fluctuating magnetic field; the spins also experience strong interactions. Figure~\ref{fig:conn} shows the Ising couplings between the spins in the January 12 Sudoku puzzle. Each spin is represented by a dot, arranged along an elliptical track for clearer visualization. A line connecting spins $m$ and $n \neq m$ indicates a nonzero coupling $J_{m, n}$. It is clear that the interactions are long-ranged, extending beyond just nearest neighbors. Closer inspection reveals that there are a total of $1,261$ connections among the $N_s = 210$ spins, and all couplings are of equal strength, with $J_{m, n} = 2$. A positive coupling constant $J$ corresponds to antiferromagnetic interactions, which energetically favor opposing spin orientations. However, satisfying the competing preferences of all these couplings simultaneously is generally impossible, a phenomenon known as frustration~\cite{Van:1977aa}. These two factors -- long-range interactions and frustration -- greatly complicate the spin optimization problem.

\begin{figure}[H]
\centering
\includegraphics[width=0.65\textwidth]{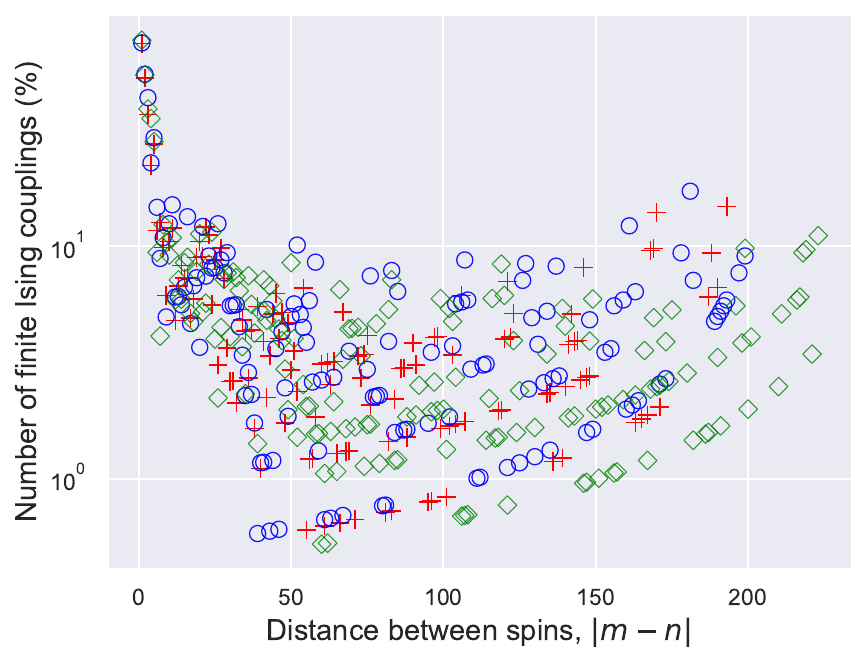}
\caption{Statistical analysis of the couplings between the spins. For nearest neighbors, about $75\%$ are coupled. The coupling fraction drops to the $10\%$ level when the distance grows to $6$ or $7$, then fluctuates between $1\%$ and $10\%$. The legends are the same as Figure~\ref{fig:hzm}. The vertical axis is in logarithmic scale.}
\label{fig:percentage}
\end{figure}

Since the long-range Ising interactions play a crucial role in our algorithm design, we separately count the number of nonzero $J_{m, n}$ values as a function of the distance $d = |m - n|$ between spins. The results are shown in Figure~\ref{fig:percentage} for the same three puzzles presented in Figure~\ref{fig:hzm}. The vertical axis, plotted on a logarithmic scale, represents the filling fraction $\rho(d)$, which is defined as the number of couplings between spins separated by distance $d$, divided by $(N_s - d)$ -- the maximum number of possible connections at distance $d$.

We observe that nearest-neighbor couplings ($d = 1$) are present in approximately $75\%$ of the possible cases. The filling fraction decreases sharply, dropping to around $10\%$ by $d = 6$, and then fluctuates between $1\%$ and $10\%$ for larger distances. Interestingly, at long distances, $\rho(d)$ begins to converge into a few smooth curves. This behavior can be traced back to the structure of the Sudoku grid and the uniform penalty terms used in the formulation.

Finally, we note that for the Sherrington-Kirkpatrick spin glass, the energy gap that separates the ground state from the first excited state is randomly generated and can be very small. This stands in contrast to the Sudoku problems where the gap is finite and known.

\section{Algorithm Design}
\label{sec:algs}


\subsection{Motivation for Designing Quantum-Inspired Algorithms}

A popular heuristic approach for solving QUBO problems is quantum annealing~\cite{finnila1994quantum, PhysRevE.58.5355, RevModPhys.80.1061}, which utilizes specialized quantum hardware, such as D-Wave's Advantage annealer~\cite{advantage}. The central idea is to engineer a time-dependent Hamiltonian, typically of the form:
\begin{displaymath}
H(t) = s(t) H_0 + \left(1 - s(t)\right) H_1,
\end{displaymath}
where $t$ represents time, measured in an appropriately chosen unit $\tau$, and the annealing schedule 
$s(t)$ is a continuous function with boundary conditions $s(t = 0) = 1$ and $s(t = 1) = 0$. At $t = 0$, the initial Hamiltonian $H(t = 0) = H_0$ typically has a known ground state $\ket{\psi_0}$, which can be prepared experimentally. $H_1$ corresponds to the QUBO problem Hamiltonian, and we aim to find its ground state $\ket{\psi_1}$. Provided that $s(t)$ varies at a sufficiently slow rate, the system's state will evolve adiabatically from $\ket{\psi_0}$ to $\ket{\psi_1}$, and the final state $\ket{\psi_1}$ can then be measured to extract the QUBO solution. For our problem, we can set $H_0 = H_x$, $H_1 = H_z$, and choose the initial state as:
\begin{equation}
\ket{\psi_0} = \ket{-}^{N_s} = \bigotimes^{N_s}_{m = 1}  \frac{1}{\sqrt{2}}\left(\ket{\uparrow}_m - \ket{\downarrow}_m\right)
\label{psi_0}
\end{equation}

In practice, several factors affect the efficacy of quantum annealing. First, during the time evolution, if the energy gap (the difference between the ground state and the first excited state) of $H(t)$ becomes vanishingly small, the time $\tau$ required for quantum annealing may grow exponentially. For finite annealing schedules, the final state will no longer be the ground state of $H_1$, and contributions from excited states could disrupt the QUBO solution. Second, a faithful implementation of 
$H(t)$ requires full connectivity between the spins. For example, the dense connection pattern in Figure~\ref{fig:conn} differs from D-Wave's Pegasus graph, where each qubit (spin) is only coupled to $15$ other qubits in a periodic pattern~\cite{advantage}. Consequently, many QUBO problems must undergo a process called embedding to fit into the annealer's hardware~\cite{advantage}. Third, quantum annealers often come with costs or restrictions on the running time, making them impractical for many users with limited resources.

These limitations motivate us to develop a quantum-inspired algorithm for solving QUBO problems. The term ``quantum-inspired'' is used in line with the review article~\cite{inspiredreview}, where it refers to algorithms that draw inspiration from quantum computation and information techniques but operate on classical data (e.g., from Sudoku or MaxCut problems) and run on classical computers rather than quantum hardware. Specifically, we map the classical problem in equation~\eqref{SGH} to a quantum spin problem in equation~\eqref{H_total} and simulate the quantum spin system classically using its tensor network representation. This approach is feasible because the low-energy quantum states of certain classes of spin Hamiltonians can be efficiently expressed as tensor networks, which can be manipulated within polynomial time~\cite{verstraete}. We will show that the quantum-inspired algorithm proposed here, sitting between classical methods such as simulated annealing and quantum hardware-based algorithms like quantum annealing and QAOA, can circumvent some of the drawbacks associated with quantum annealing. 


\subsection{A Quantum-Inspired Algorithm}
\label{subsec:alg}


A key component of our quantum-inspired algorithm is the tensor network representation of many-spin wave functions~\cite{verstraete, tensorreview}. In particular, we leverage the Matrix Product States (MPS) formalism to efficiently represent and manipulate quantum spin states on classical computers. Existing tensor network approaches to QUBO problems---such as those recently proposed by Patra et al.~\cite{peps} and Mugel et al.~\cite{portfolio}---typically aim to find ground states of Ising spin glass models via imaginary-time evolution. However, this method can be computationally expensive, especially for systems with long-range interactions.

To address this challenge and better guide the MPS toward the ground state, we adopt an alternative approach based on the Density Matrix Renormalization Group (DMRG)~\cite{white,dmrgreview}. DMRG is an iterative variational algorithm that optimizes MPS representations by ``sweeping'' through the spin chain multiple times to progressively minimize the system's energy. A brief introduction to MPS and DMRG, along with relevant references, is provided in Appendix~\ref{appendix:A}.

In addition to MPS and DMRG, the third ingredient of our algorithm is the driver Hamiltonian, a concept borrowed from quantum annealing (analogous to the mixer Hamiltonian in QAOA). Instead of using the traditional, slow, continuous annealing schedule $s(t)$ employed in quantum annealing, we adopt a discrete $M$-step driving schedule to steer the system toward the ground state.

\begin{displaymath}
H_x = H_1 \rightarrow H_2 \rightarrow \cdots \rightarrow H_M = H_z
\end{displaymath}

The Hamiltonian at the $i$-th step is given by
\begin{align*}
H_i = a_i H_x + b_i H_z, & & i = 1, 2, \ldots, M,
\end{align*}
where the driving parameters $a_i$ and $b_i$ control how much the problem Hamiltonian $H_z$ and the driver Hamiltonian $H_x$ are mixed at the $i$-th step. And the entire set $\{a_i, b_i\}$ specifies a discrete driving path. For the starting and ending point, we can set $a_1 = 1, b_1 = 0$ and $a_M = 0, b_M = 1$.

The algorithm proceeds as described in Algorithm~\ref{alg:core}.

\begin{algorithm}[H]
\caption{Quantum-inspired algorithm}
\begin{algorithmic}[1]
\STATE Choose an initial matrix product state $\ket{\psi_0}$.

\FOR{$i = 1$ to $M$}

\STATE update the MPS for $H_i$ from $\ket{\psi_{i - 1}}$ to $\ket{\psi_i}$ by DMRG sweeps,
\begin{displaymath}
\ket{\psi_{i - 1}} \xrightarrow{\text{DMRG sweeps for $H_i$}} \ket{\psi_i}
\end{displaymath}

\ENDFOR

\RETURN $\ket{\psi_M}$
\end{algorithmic}
\label{alg:core}
\end{algorithm}

Once we obtain $\ket{\psi_M}$ from Algorithm~\ref{alg:core}, the ground state of the Ising spin glass $H_z$, we can compute both the ground energy $E_M$ and the corresponding spin configuration. We shall show in Section~\ref{sec:experimentS} and Section~\ref{sec:experimentM} below that the combination of MPS, DMRG and discrete driving provides a more tractable and efficient approach for QUBO problems.

One might wonder why we don't simply use DMRG to find the ground state of $H_z$ directly, thereby bypassing all the preceding steps up to $i = M - 1$. The reason is that DMRG rarely works effectively without a good initial guess for the starting state. While DMRG is an excellent tool for finding the ground state of one-dimensional gapped Hamiltonians with local interactions, it can easily get stuck in local minima when applied to glassy or long-range Hamiltonians, such as $H_z$. In our algorithm, all the preceding steps are designed to create a pathway that successively steers the MPS state:
\begin{displaymath}
\ket{\psi_0} \rightarrow \ket{\psi_1} \rightarrow \cdots \rightarrow \ket{\psi_M},
\end{displaymath}
such that $\ket{\psi_{M - 1}}$ is sufficiently close to the ground state of $H_z$. The success of this heuristic approach depends on making careful choices for both the initial state $\ket{\psi_0}$ and the driving path $\{H_i\}$, tailored to the specific problem at hand. An analogy for MPS-DMRG-based search is given in Appendix~\ref{appendix:B}. 

Our algorithm is implemented in Julia 1.10.5, using the ITensors library version 0.6.23 to carry out the MPS and DMRG calculations~\cite{itensor}. All runtime data were collected on a laptop equipped with an Apple M3 chip. 

The discrete driving schedule proposed here bears some resemblance to the concept of `fictitious time' used in quantum annealing via path-integral Monte Carlo~\cite{Santoro2002, pimc2002}, where a transverse field is decreased over a series of discrete Monte Carlo steps. However, it is important to note that the number of Monte Carlo steps in such methods is typically very large (up to $10^6$)~\cite{pimc2002}, whereas the number of driving steps employed in our work is comparatively small -- on the order of $10$.


\section{Solving Sudoku Puzzles}
\label{sec:experimentS}


In this section, we test the performance of our algorithm by applying it to solve the Sudoku puzzles published daily by The New York Times~\cite{nycom}. This choice is inspired by an example presented by Mücke~\cite{mucke}. A key advantage of using these puzzles is that they are archived and readily accessible online~\cite{nynet}.

Table~\ref{tb:NYT} presents a few examples. Each row lists a puzzle, uniquely identified by its date of publication and assigned difficulty level, where ``$m$'' stands for intermediate and ``$h$'' for hard. Also displayed for each puzzle are the number of clues ($N_c$), the number of up spins in the correct solution ($N_{\uparrow}$), the total number of spins ($N_s$) in the corresponding Ising spin glass $H_z$, and the constant energy shift $c_1 + c_2 + c_3$ in $H_z$ (see~\eqref{SGH}). A successful run of the algorithm should drive the energy down to zero. For verification, we take the converged spin configuration, translate it back to the $X$ variables, then to the full QUBO vector $Z$, and print the completed Sudoku board to ensure that all numbers fit correctly and none of the constraints are violated.

\begin{table}[H]
\caption{A few examples of solved Sudoku puzzles in chronological order according to their publication date in The New York Times. These puzzles vary in the number of clues ($N_c$), total number of spins ($N_s$), and other parameters; see the main text for details.}
\label{tb:NYT}
\newcolumntype{L}{>{\raggedright\arraybackslash}X}
\newcolumntype{C}{>{\centering\arraybackslash}X}
\resizebox{1\textwidth}{!}{\begin{tabular}{p{3cm}||p{2cm}|p{2cm}|p{2cm}|p{2cm}|p{2cm}}
\specialrule{\heavyrulewidth}{0pt}{0pt}
Puzzle date & Level & $N_c$ & $N_{\uparrow}$ & $N_s$ & $c_1 + c_2 + c_3$\\ \hline
January 8, 2024 & $h$   & 24    & 57 &   211 & 368.5 \\
December 10, 2024   & $m$  & 26    & 55 &   200 & 359.0\\
January 2, 2025   & $m$ & 22    & 59 &   250 & 520.5\\
January 8, 2025   & $m$  & 23    & 58 &   232 & 461.5\\
January 12, 2025   & $m$ & 27    & 54 &   210 & 426.5\\
January 14, 2025   & $m$  & 23    & 58 &   232 & 457.0\\
January 15, 2025   & $m$  & 25    & 56 &   210 & 380.0\\ 
\specialrule{\heavyrulewidth}{0pt}{0pt}
\end{tabular}}
\end{table}

We discuss a few examples in some detail to illustrate how the algorithm parameters are chosen for the calculations. For the January 8, 2024 puzzle, listed in the first row of the table, we start from the state~\eqref{psi_0}, which is an equal-weight superposition of all $S^z$ basis states, and use a homogeneous transverse field $h^x_m = 0.7$ in the driver Hamiltonian $H_x$. The driving schedule is linear.
\begin{align*}
b_i = \frac{i}{M}, & & a_i = 1 - b_i, & & M = 10
\end{align*}

For DMRG, we set the maximum bond dimension $D = 60$ and perform $5$ sweeps at each stop $i$. With these settings, the proposed algorithm successfully solves the puzzle. Both $H_i$ and $\ket{\psi_i}$ change with $i$, so the time spent on each DMRG sweep varies roughly from $0.1$ to $8$ seconds. In fact, the algorithm spends most of its time on $H_{i < M}$ to steer the MPS. During these steps, the MPS have $D = 60$, which makes them computationally expensive. Only in the last step ($i = M$) is the problem Hamiltonian $H_z$ considered. This final stage is efficient and fast: the MPS typically converges within $3$ DMRG sweeps, and the bond dimension drops from $60$ to $1$, meaning the ground state becomes a product state, as expected. The quick convergence is largely due to the proximity of $\ket{\psi_{M - 1}}$ from the preceding steps to the true ground state $\ket{\psi_M}$.

Both the initial state and the driving schedule used here resemble those in quantum annealing. This is intentional, allowing us to take advantage of superposed states and the mixer Hamiltonian. However, the similarity ends there. Our algorithm does not follow any adiabatic time evolution, so the ``trajectory'' of the MPS wave function from $\ket{\psi_0}$ to $\ket{\psi_M}$ is fundamentally different. The algorithm is largely insensitive to the choice of $\ket{\psi_0}$, as long as it serves as a reasonable starting point for the DMRG solution of $H_1$, which is predominantly $H_x$ with only a small admixture of $H_z$.

\begin{figure}[H]
\centering
\includegraphics[width=0.55\textwidth]{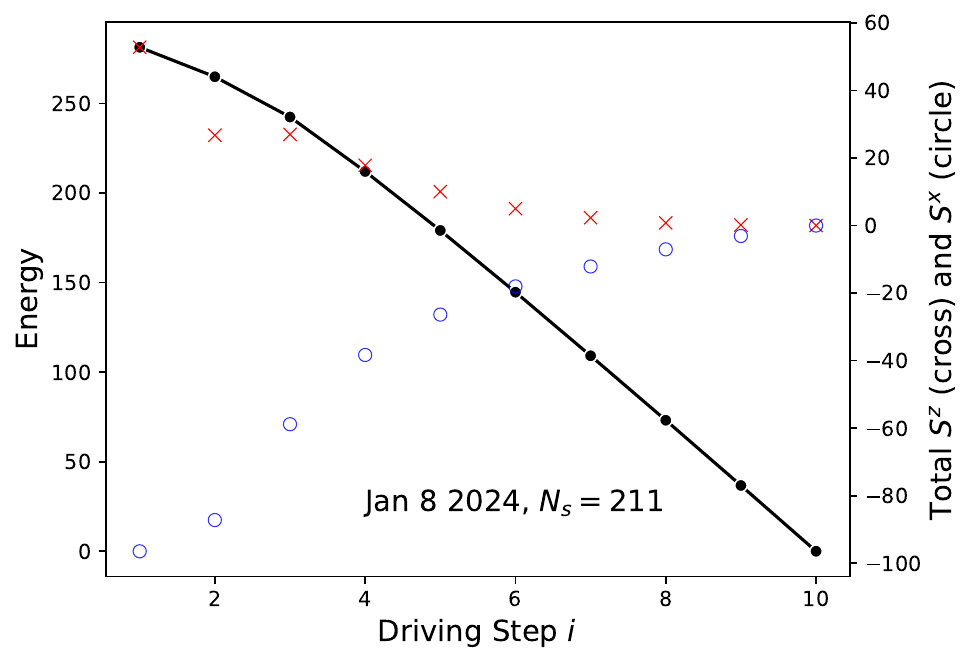}
\caption{How the January 8, 2024 puzzle gets solved: As the driving field $h_x$ is reduced, the energy gradually drops to zero. The total spin along the $x$-direction vanishes, while the total spin along the 
$z$-direction approaches its ground state value. The bond dimension is $D = 20$ and the driving field is set to $h_x = 0.75$.}
\label{fig:mucke}
\end{figure}

To demonstrate the flexibility of the algorithm, we use the same puzzle but start from a random MPS with bond dimension $3$ and choose $D = 20$ for all subsequent steps. The puzzle is solved successfully at a much faster pace, with each sweep taking about $0.8$ seconds. As shown in Figure~\ref{fig:mucke}, the energy (black dots) steadily drops to zero. To track the spins' behavior at each step, we define the total spin-$z$ operator measured from its ground state value $\left({N_{\uparrow} - \frac{N_s}{2}}\right)$,
\begin{equation}
{S^z} = \sum^{N_s}_{m = 1} {S^z_m} - \left({N_{\uparrow} - \frac{N_s}{2}}\right)
\end{equation}
and evaluate its expectation value at the current MPS state $\ket{\psi_i}$.

The red crosses in Figure~\ref{fig:mucke} show the expectation values of ${S^z}$ (magnified by a factor of $4$ to be clearly visible). Figure~\ref{fig:mucke} also includes the expectation values (blue circles) of the total spin-$x$ operator defined by 
\begin{equation}
{S^x} = \sum^{N_s}_{m = 1}  {S^x_m}
\end{equation}

As the weight of $H_x$ is decreased and the weight of $H_z$ is increased during the driving process, ${S_x}$ gradually approaches zero, while ${S_z}$ converges to its ground state value ${N_{\uparrow} - \frac{N_s}{2}}$. An interesting period occurs between driving steps $2$ and $4$. During this interval, the spins begin to transition from being predominantly aligned along the $x$-axis to aligning along the $z$-axis. The inflection point at this transition appears crucial for the success of the algorithm.

If the primary concern is the wall time for solving the $9 \times 9$ Sudoku puzzles, our algorithm pales in comparison to classical algorithms that build on logic and exploit the structure of the problem. In contrast, the QUBO approach to Sudoku puzzles, as noted earlier, is not efficient and resembles more of a blind search. The problem is treated simply as another QUBO matrix, and the solution strategy here does not focus on the underlying meaning of the original constraints. However, this apparent weakness can be turned into a strength. As the board size increases, the algorithm is expected to outperform conventional methods since the calculation scales as $N_s D^3 M$. More importantly, it can be applied to other QUBO problems, as we will demonstrate in Section~\ref{sec:experimentM}. 

Other puzzles listed in the table are solved in a similar manner. Since they differ in $N_s$, $h^z_i$, and $J_{i, j}$, directly copying the algorithm parameters from a previously successful example may not always work. Occasionally, the final state might become trapped in a local minimum, with energy higher than zero (indicating that some constraints are violated). In such cases, the solution is simple: restart from a different MPS state. A useful strategy to ``shake'' the system loose and escape local minima is to tune the driving term $H_x$, for example, by adding a random fluctuation to the transverse field.
\begin{equation}
h^x_m = h_x + \eta_m
\end{equation}
where $h_x$ is the homogeneous part and $\eta_m$ is a random variable at site $m$ drawn from a uniform distribution within the interval $(-\eta, \eta)$. For example, in solving the January 15, 2025 puzzle in the last row of the table, we set $D = 20$, $h_x = 1.3$, and $\eta = 0.2$ at each driving step. The algorithm once again succeeds in $10$ steps.

\begin{figure}[H]
\centering
\includegraphics[width=0.55\textwidth]{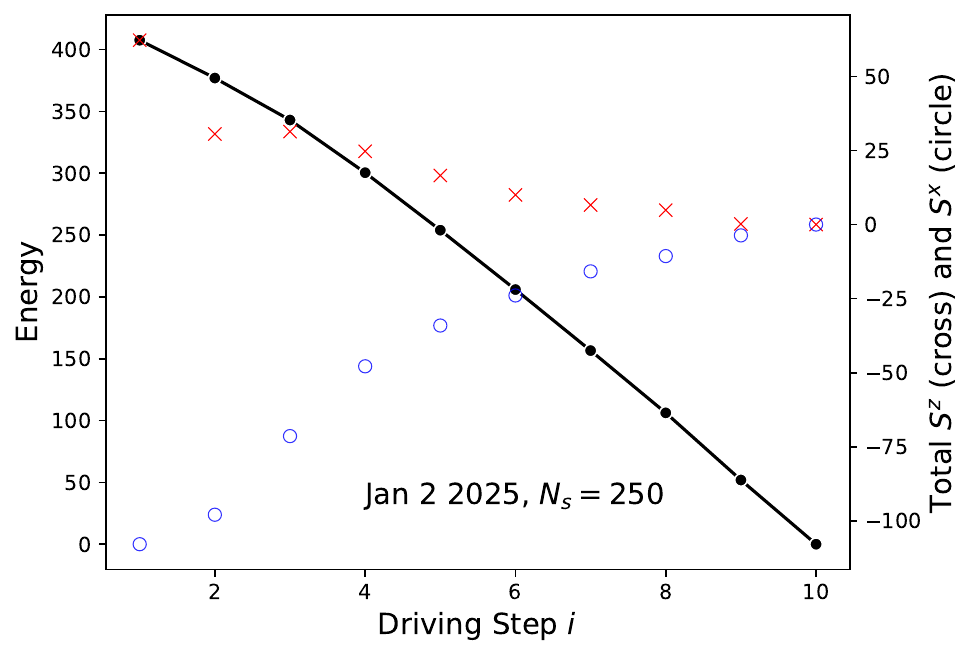}
\caption{The solution process for another puzzle with fewer clues and more spins: $D = 20$, $h_x = 1$.}
\label{fig:jan2E}
\end{figure}

\begin{figure}[H]
\centering
\includegraphics[width=0.5\textwidth]{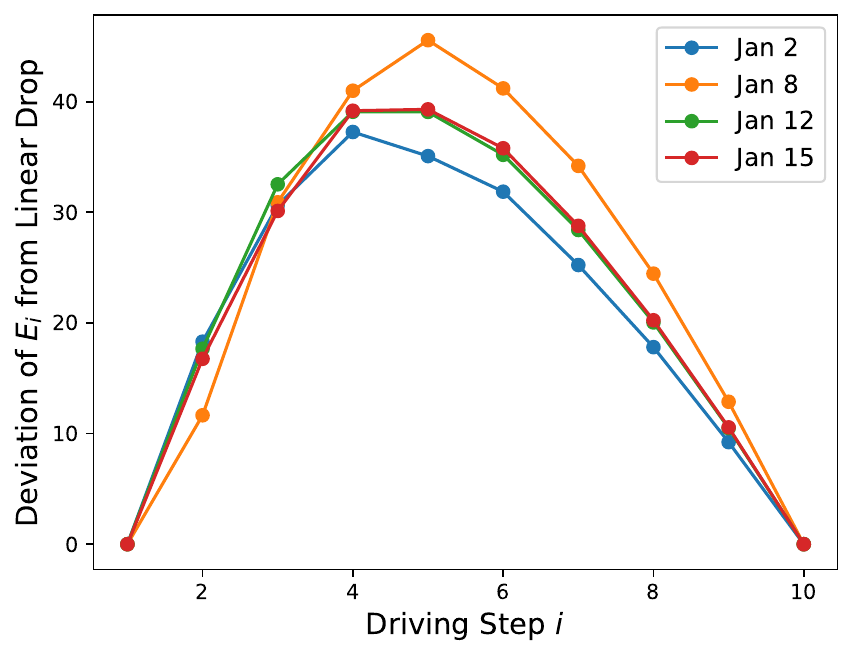}
\caption{The nonlinear drop in energy during the driving process. See main text for details.}
\label{fig:devi}
\end{figure}

\begin{figure}[H]
\centering
\includegraphics[width=0.6\textwidth]{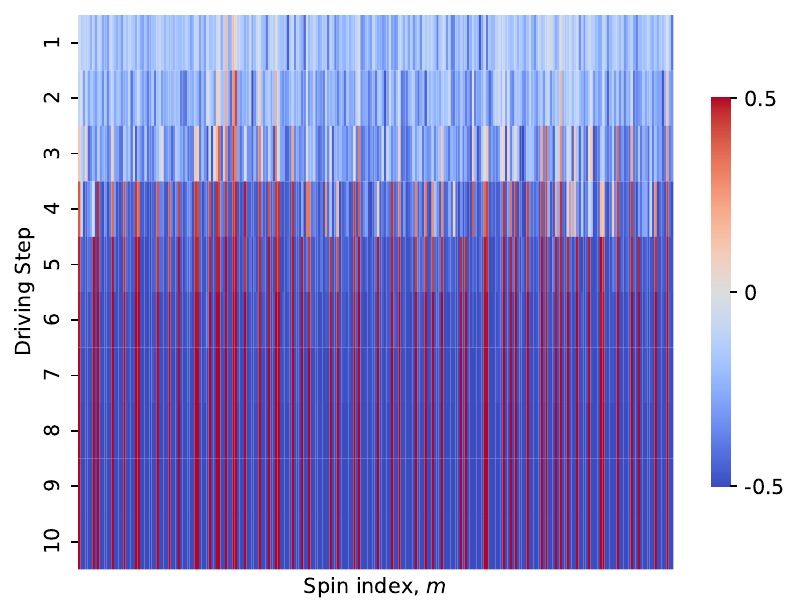}
\caption{The evolution of the 250 spins during the solution process of the January 2, 2025 puzzle. The value of ${S^z_m}$ is color-coded: red represents spin-up, and blue represents spin-down.}
\label{fig:heatmap}
\end{figure}

Within the table, the January 2, 2025 puzzle has the least number of clues and the maximum number of spins, $N_s = 250$. The broad features of its solution process can be appreciated from Figure~\ref{fig:jan2E}, which is comparable to our earlier example in Figure~\ref{fig:mucke}. In both cases, the energy decreases with $i$, but not exactly in a linear fashion. To quantify this, let $E_i$ be the energy achieved at the $i$th driving step. If we draw a straight line from the starting point $(i = 1, E_1)$ to the ending point $(i = M, E_M)$, then let $\delta E_i$ be the deviation of $E_i$ from this line. We plot $\delta E_i$ in Figure~\ref{fig:devi} for a few solved puzzles. The energy drop is clearly sub-linear, and the peak location of $\delta E_i$ correlates with the inflection point of the total ${S^x}$. A more detailed picture of the solution process is provided in Figure~\ref{fig:heatmap}. Here, the horizontal axis represents the spin index $m = 1$ to $250$, going from left to right. The vertical axis corresponds to the driving step $i = 1$ to $M = 10$, going from top to bottom. On each row, shown in false color, are the expectation values of the spin-$z$ operator, ${S^z_m}$, computed at the corresponding step.

We observe that in the initial steps, due to the $h_x$ field, ${S^z_m}$ appears quite random. By step 4, the spins begin to settle into their ``correct'' states, i.e., the expectation values of $S^z_m$ align with the final solution. Once again, we see that the first few driving steps are crucial for the algorithm to work. However, note that the MPS wave function at this stage still has a large $D = 20$, and the energy remains relatively high. In the last step, after only three DMRG sweeps, the correct solution emerges as a product state, with red and blue colors representing $\ket{\uparrow}$ and $\ket{\downarrow}$, respectively. The spin configuration is read out and converted back first to the QUBO vector $X$, and then to the original binary variables in $Z$, resulting in the solved Sudoku board (with the clues in blue).
\setlength{\tabcolsep}{4.6pt}
\begin{center}
\begin{tabular}{|l|l|l|l|l|l|l|l|l|}
\hline
 2  & \textcolor{blue}{8} & 1 & \textcolor{blue}{3} & 5 & 4 & 6 & 7 & \textcolor{blue}{9}   \\ \hline
 \textcolor{blue}{9} & 7 & \textcolor{blue}{3} & 8 & 1&  6 & \textcolor{blue}{5} & 2 & 4  \\ \hline
5 & \textcolor{blue}{6} & 4 & 7 & 9 & \textcolor{blue}{2} & 8 & \textcolor{blue}{1} & 3  \\ \hline
  \textcolor{blue}{1} & 3 &  9 & \textcolor{blue}{4} & 6 & 8 & 2 & \textcolor{blue}{5} & 7  \\ \hline
 8  &2  & 5 & \textcolor{blue}{9} & 7 & 3 & 1 & 4 & 6  \\ \hline
 \textcolor{blue}{6}  & 4 & 7 & 5  & 2 & 1 & 9 & 3 & \textcolor{blue}{8}  \\ \hline
 \textcolor{blue}{4} & 9 & 2 & 6 & 3  & 5 & 7 & 8 & \textcolor{blue}{1}  \\ \hline
 3 & 5 & \textcolor{blue}{6}&  1 & \textcolor{blue}{8}  &\textcolor{blue}{7} & 4 & 9  &\textcolor{blue}{2}  \\ \hline
 7 & 1 & 8  & 2  & 4  & \textcolor{blue}{9} & 3 & 6  &5  \\ \hline
\end{tabular}
\end{center}

Interestingly, this valid solution differs from the one offered in~\cite{nynet}. Our quantum-inspired algorithm has found an independent solution!

We conclude the case study of Sudoku with a final remark. Our algorithm is heuristic and relies on several tuning parameters. It is unrealistic to expect a universal set of fail-safe parameters that can solve all Sudoku puzzles, as they translate into very different spin Hamiltonians. Therefore, some trial and error may be required when tackling a new puzzle. Compared to hardware-based quantum algorithms, a major advantage of our quantum-inspired approach is the explicit knowledge of the MPS wave functions, which allow for a detailed diagnosis of the solution process. As demonstrated above, in Figures~\ref{fig:mucke} through~\ref{fig:heatmap}, this analysis helps us fine-tune the initial state, driver, driving schedule, and other parameters. These strategies for parameter optimization will be further discussed when the algorithm is applied to MaxCut problems.


\section{Solving the MaxCut Problems}
\label{sec:experimentM}


In this section, we apply the proposed algorithm to MaxCut, one of the most well-known problems in combinatorial optimization~\cite{Karp1972}. The goals are twofold: First, by testing it on new problem instances, we demonstrate the versatility of the algorithm and its capability to solve QUBO problems with varying characteristics. Second, by analyzing its performance across different problem sizes and types, we illustrate how the algorithm can be fine-tuned to optimize its performance for specific instances.

The MaxCut problem is defined on an undirected graph $G = (V, E)$, where $V$ represents a set of vertices and $E$ represents a set of edges. An edge connecting vertex $i$ and vertex $j$ is denoted by $(i, j) \in E$ and is assigned a weight $w_{i, j}$. The objective is to partition $V$ into two sets, $V = A \cup \bar{A}$ with $A \cap \bar{A} = \emptyset$, such that the weighted sum of the edges connecting $A$ to $\bar{A}$ is maximized. This problem is NP-hard, and its connection to spin glass Hamiltonians has long been recognized~\cite{Karp1972, spglassmaxcut}.

To cast the problem into QUBO form, let us define the binary variable $x_i = 1$ if vertex $i$ belongs to set $A$, and $x_i = 0$ otherwise. The edge $(i, j)$ is part of the cut if and only if $x_i \neq x_j$, i.e., $(x_i - x_j)^2 = 1$. Therefore, the objective is to maximize the cut value.

\begin{displaymath}
\sum_{(i, j) \in E} w_{i, j} (x_i - x_j)^2    
\end{displaymath}
or equivalently to minimize the cost function
\begin{align*}
\min & f(\{x_i\}) = \sum_{(i, j) \in E} w_{i, j} (2 x_i x_j  - x_i^2 - x_j^2)
\end{align*}

It is easy to cast $f$ into the standard QUBO form
\begin{displaymath}
f = \sum_{i, j} x_i Q_{i, j} x_j
\end{displaymath}
We then map the MaxCut problem to an Ising spin glass Hamiltonian similar to~\eqref{SGH}, but with $c_1 = c_2 = 0$. In this case, the cost function depends on the weight matrix $w$, and there is no penalty from constraints. This setup is almost the opposite of the Sudoku case, where the cost function is entirely derived from the constraints. Despite these differences, both problem types ultimately lead to spin Hamiltonians of the general form~\eqref{SGH}, which we solve by adding a driver term $H_x$ and designing an appropriate driving schedule.

We evaluate the performance of the algorithm using MaxCut instances drawn from the Biq Mac Library. All instances used in this section are available for download at~\cite{biqmac} or alternatively at~\cite{instances}. We will refer to each instance by its file name, which contains $N_V$, the number of vertices. Note that the complexity of the problem also depends crucially on $N_E$, the number of edges, as well as the values of $w_{i, j}$. In what follows, we discuss three batches of test cases, labeled (I) to (III), with increasing difficulty. Each batch has a distinctive distribution of edge weights.

\begin{table}[H]
\caption{Twenty weighted MaxCut instances and the parameters used to solve them. Each instance is identified by its file name in the Biq Mac Library. The edge weights are either $1$ or $-1$. For the first ten instances, the number of vertices $N_V = 80$ and the number of edges $N_E = 316$. For the next ten instances, $N_V = 100$ and $N_E = 495$. The algorithm successfully finds $E_0$, the ground state energy of the spin model (the MaxCut value is simply $-E_0$), except for the case pm1s\_100.5.}
\label{tb:weighted}
\begin{tabularx}{\textwidth}{
    >{\raggedright\arraybackslash}X||  
    >{\centering\arraybackslash}X|    
    >{\centering\arraybackslash}X|
    >{\centering\arraybackslash}X|
    >{\centering\arraybackslash}X|
    >{\centering\arraybackslash}X
}
\specialrule{\heavyrulewidth}{0pt}{0pt}
Instance & $E_0$ & $M$ & $h_x$ & $\eta$ & $D$\\
\hline
pm1s\_80.0   & $-$79   & 5    & 1&   0 &  30 \\
pm1s\_80.1   & $-$69 & 5    & 1&   0 & 30\\
pm1s\_80.2   & $-$67  & 5    & 1&  0.3 & 30\\
pm1s\_80.3   & $-$66 & 5    & 1&   0 & 30\\
pm1s\_80.4   & $-$69  & 5    & 1&   0.3 &30\\
pm1s\_80.5   & $-$66 & 20    & 1&   0.3 & 30\\
pm1s\_80.6  & $-$71  & 5    & 1&   0 & 30\\
pm1s\_80.7   & $-$69  & 5    & 1&   0 & 30\\
pm1s\_80.8   & $-$68  & 5    & 1&   0 & 30\\
pm1s\_80.9   & $-$67  & 5    & 1&   0 & 30\\
\hline
pm1s\_100.0   & $-$127   & 5    & 1&   0.3 &  30 \\
pm1s\_100.1   & $-$126 & 5    & 1&   0.3 & 30\\
pm1s\_100.2   & $-$125  & 10    & 1&  0.3 & 30\\
pm1s\_100.3   & $-$111 & 10  & 1&   0.3 & 30\\
pm1s\_100.4   & $-$128  & 5    & 1&   0.3 &30\\
pm1s\_100.5   & $-$\underline{125} & 10    & 1&   0 & 60\\
pm1s\_100.6  & $-$122  & 10    & 1&   0.3 & 30\\
pm1s\_100.7   & $-$112  & 20    & 1&   0.3 & 60\\
pm1s\_100.8   & $-$120  & 10   & 1&   0.3 & 30\\
pm1s\_100.9   & $-$127  & 5    & 1&   0.3 & 30\\
\specialrule{\heavyrulewidth}{0pt}{0pt}
\end{tabularx}
\end{table}


\paragraph{I: The first batch of instances listed in Table~\ref{tb:weighted}.}

\begin{figure}[H]
    \centering
    \begin{subfigure}[b]{0.5\textwidth}
        \includegraphics[width=\textwidth]{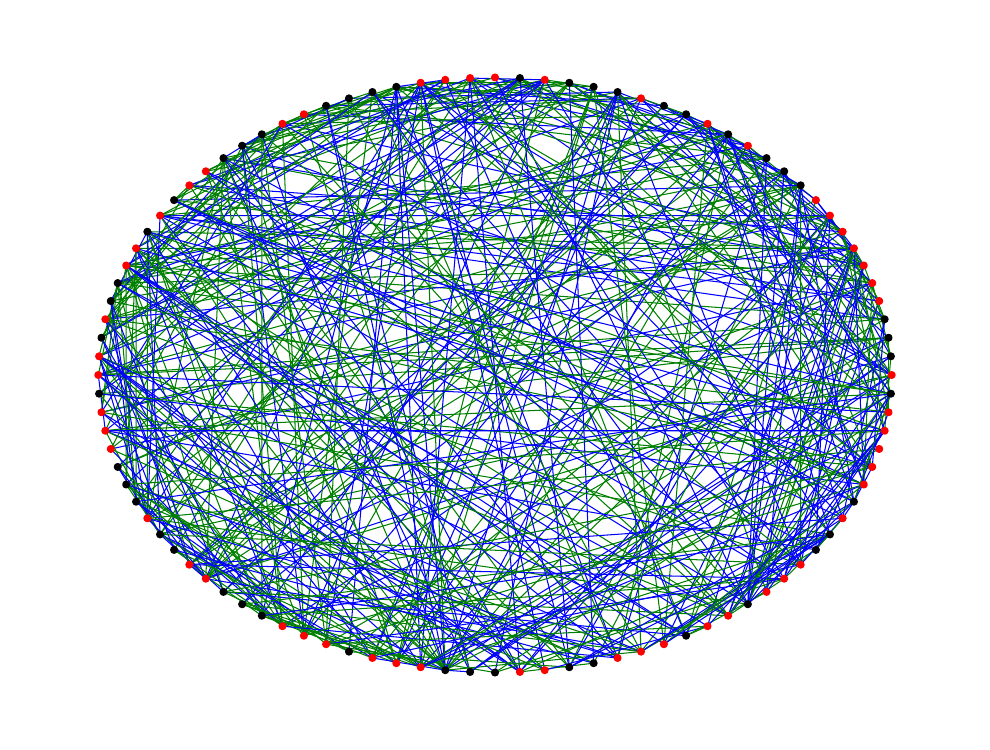}
        \subcaption{}
        \label{fig:left}
    \end{subfigure}%
    \hfill
    \begin{subfigure}[b]{0.5\textwidth}
        \includegraphics[width=\textwidth]{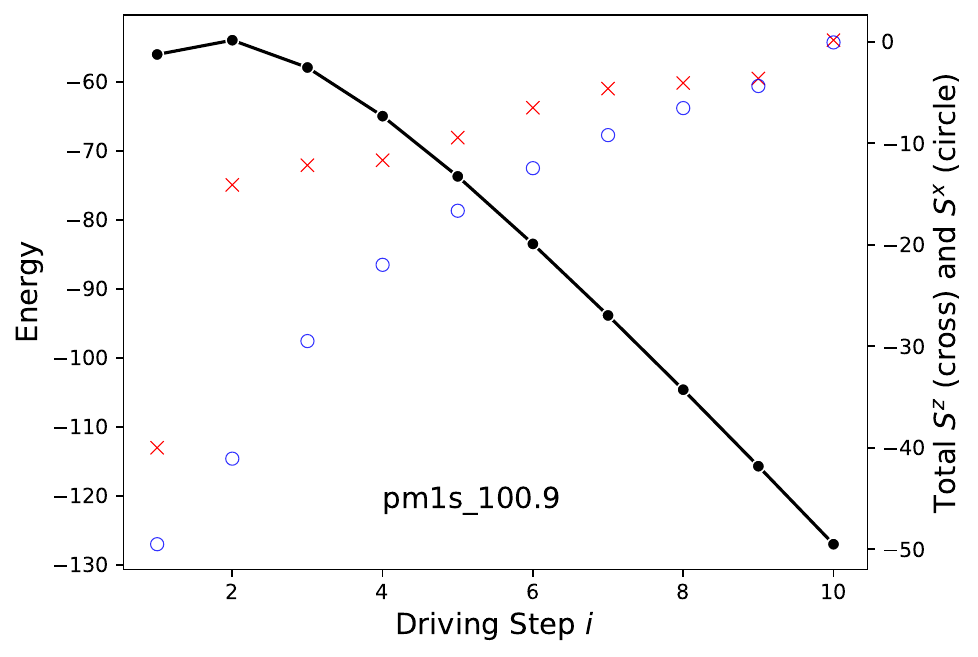}
        \subcaption{}
        \label{fig:right}
    \end{subfigure}
    \caption{(\textbf{a}) Graph representation of the problem. The edges carry weights of $1$ (in green) or $-1$ (in blue). The solution is indicated by the node color: nodes in black belong to set $A$, while the rest of the nodes (in red) belong to set $\bar{A}$. (\textbf{b}) The energy (dot), expectation values of ${S^z}$ (cross, measured from its ground state value and magnified by $10$ times) and ${S^x}$ (circle) during the driving process. Solving the weighted MaxCut instance \texttt{pm1s\_100.9} from the Biq Mac Library.}
    \label{fig:mc9}
\end{figure}

These are weighted MaxCut problems where the edge weight $w_{i, j}$ is set to either $1$ or $-1$. An example is illustrated in Figure~\ref{fig:left}, which features $100$ nodes and $495$ edges. The edges are color-coded: green for $w_{i, j} = 1$ and blue for $w_{i, j} = -1$. It is helpful to compare Figure~\ref{fig:mc9} with the Sudoku example in Figure~\ref{fig:conn}, where the edge weight is constant. In the MaxCut case, the edges are not only denser but also carry signs. In the spin language, this means there are more couplings among the Ising spins, and the coupling can be either ferromagnetic or antiferromagnetic. Also, the magnetic field along the $z$-axis for each spin $m$ vanishes in this case, i.e., $h^z_m = 0$.


Despite these differences, our algorithm successfully solves all instances listed in Table~\ref{tb:weighted}, yielding the correct MaxCut values, with one exception: the instance named pm1s\_100.5. In this case, the lowest energy obtained, $E_0 = -125$, is above the value of $-128$ quoted in~\cite{biqmac,instances}. To alert the reader, the $E_0$ value is underscored to indicate that this case remains unresolved. The parameters used in each instance (number of driving steps $M$, transverse field $h_x$ and its random fluctuations $\eta$, and bond dimension $D$) are listed in Table~\ref{tb:weighted}. Five DMRG sweeps are used for each driving step, with one exception: for pm1s\_100.6, 10 sweeps are used to improve accuracy.

Note that the algorithm parameters listed in the tables in this section may not be optimal. The problem can sometimes be solved with fewer resources than listed here.

Comparing the rows of Table~\ref{tb:weighted} reveals some strategies for solving new QUBO problems. Within each group of MaxCut instances of similar size, some instances (e.g., pm1s\_80.5 and pm1s\_100.7) are more challenging to optimize. It is also generally harder to reach the ground state for instances with larger $N_V$ and $N_E$. To tackle these harder cases, increasing the driving steps $M$ and shaking the system with random fluctuations $\eta$ often helps. Sometimes, increasing the bond dimension (e.g., $D = 60$ for pm1s\_100.7) is necessary. Typically, these tuning decisions are made by monitoring the convergence of energy during the driving process. One example is shown in Figure~\ref{fig:right} for the instance pm1s\_100.9.

\paragraph{II. The second batch of instances listed in Table~\ref{tb:unweighted}.}

\begin{figure}[H]
    \centering
    \begin{subfigure}[b]{0.49\textwidth}
        \includegraphics[width=\textwidth]{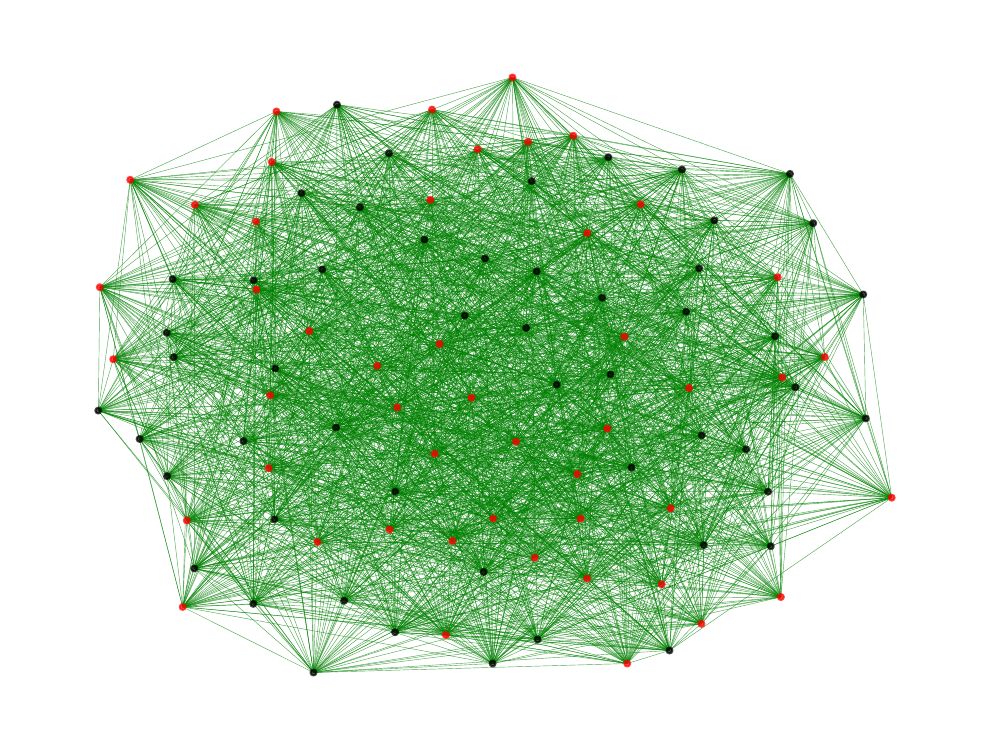}
        \subcaption{}
        \label{fig:newleft}
    \end{subfigure}
    \hfill
    \begin{subfigure}[b]{0.49\textwidth}
        \includegraphics[width=\textwidth]{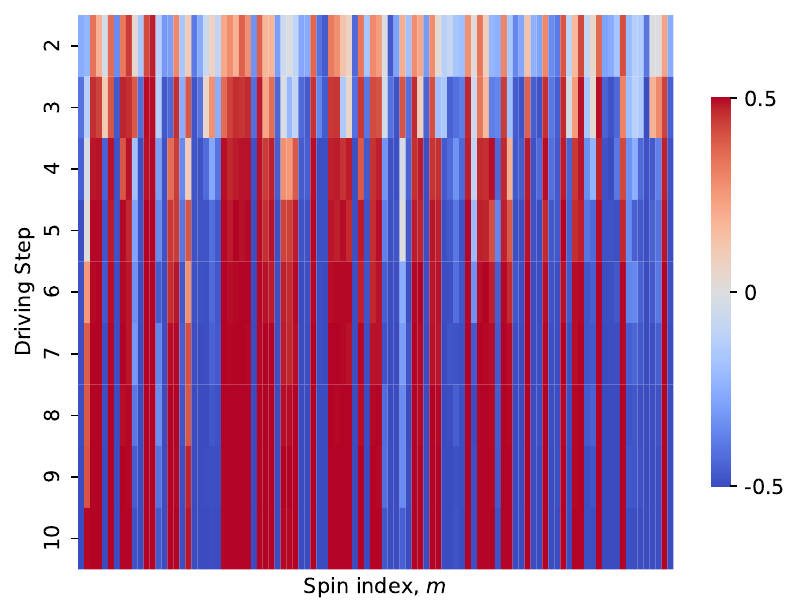}
        \subcaption{}
        \label{fig:newright}
    \end{subfigure}
    \caption{(\textbf{a}) Graph representation of the problem. All edges (green lines) carry equal weight of 1. The MaxCut found by our algorithm is color-coded: nodes in black belong to set $A$, while nodes in red belong to set $\bar{A}$. (\textbf{b}) The expectation value of $S^m_z$ during the driving process as the spins gradually settle into their ground state orientation (red represents spin up, blue represents spin down). Only driving steps $2$ to $10$ are shown to enhance the color contrast. Solving the unweighted MaxCut instance \texttt{g05\_100.4} from the Biq Mac Library.}
    \label{fig:mc4}
\end{figure}

These are unweighted MaxCut problems where all the edges share the same weight of $1$, but the connections are much denser than in the first batch. For instance, the problem shown in Figure~\ref{fig:newleft} has the same number of nodes as in Figure~\ref{fig:mc9}, but the number of edges has increased fivefold to $2475$. The spin-spin coupling is so dense that it may seem challenging for MPS and DMRG, as they are typically known to work well with spin Hamiltonians involving local interactions. Amazingly, the algorithm continues to yield correct MaxCut values with only a moderate increase in the bond dimension $D$, despite the steep rise in $N_E$.


The heatmap of $S^m_z$ in Figure~\ref{fig:newright} provides further insight into the solution process for the instance g05\_100.4. Regarding the choice of algorithm parameters, we observe similar trends noted earlier from the first batch in Table~\ref{tb:weighted}: for larger $N_V$ and $N_E$, we need to increase the number of driving steps or the number of DMRG sweeps to steer the system toward the ground state. Occasionally, the bond dimension has to be increased (e.g., $D = 60$ for g05\_60.7 and g05\_100.8).

Note that there are two cases in Table~\ref{tb:unweighted} (with the $E_0$ values underscored) where the algorithm came very close to the ground state but did not quite reach it.

\begin{table}[H]
\caption{Twenty unweighted MaxCut instances from the Biq Mac Library with dense edges. The first ten cases have $N_V = 60$ and $N_E = 885$, while the remaining cases have $N_V = 100$ and $N_E = 2,475$. All edge weights are set to $1$. The algorithm successfully solves all instances except g05\_100.3 and g05\_100.5, where the $E_0$ values should be $-1,424$ and $-1,436$, respectively~\cite{biqmac, instances}. Here, $N_{sw}$ refers to the number of DMRG sweeps per driving step, with $h_x = 1$ and $\eta = 0.3$.}
\label{tb:unweighted}
\newcolumntype{C}{>{\centering\arraybackslash}X}
\begin{tabularx}{\textwidth}{
    >{\raggedright\arraybackslash}X||  
    >{\centering\arraybackslash}X|    
    >{\centering\arraybackslash}X|
    >{\centering\arraybackslash}X|
    >{\centering\arraybackslash}X   
}
\specialrule{\heavyrulewidth}{0pt}{0pt}
Instance & $E_0$ & $M$ & $N_{sw}$ & $D$\\
 \hline
g05\_60.0   & $-$536   & 10    &5 &  30 \\
g05\_60.1    & $-$532 & 10   &  5& 40\\
g05\_60.2    & $-$529  & 10    &  5 & 30\\
g05\_60.3    & $-$538 & 10  &   5& 30\\
g05\_60.4    & $-$527  & 10   &  5 &30\\
g05\_60.5    & $-$533 & 10    &  5 & 30\\
g05\_60.6   & $-$531  & 10    &  5 & 30\\
g05\_60.7    & $-$535  & 10    &  5 & 60\\
g05\_60.8    & $-$530  & 10   &  5 & 30\\
g05\_60.9    &$-$533  & 20   &  5 & 40\\
 \hline \hline
g05\_100.0   & $-$1430   & 20    &5 &  40 \\
g05\_100.1    & $-$1425 & 20   &  5& 40\\
g05\_100.2    & $-$1432 & 20    & 5& 40\\
g05\_100.3    & $-$\underline{1423} & 10  & 5 & 30\\
g05\_100.4    & $-$1440  & 20   & 5 &40\\
g05\_100.5    & $-$\underline{1435} & 10    & 5 & 30\\
g05\_100.6   & $-$1434  & 10    &10  & 40\\
g05\_100.7    & $-$1431  & 10    & 10& 40\\
g05\_100.8    & $-$1432  & 10   &  10& 60\\
g05\_100.9    &$-$1430  & 10   &  10& 40\\
\specialrule{\heavyrulewidth}{0pt}{0pt}
\end{tabularx}
\end{table}


\paragraph{III. The third batch of instances listed in Table~\ref{tb:bqp}.}

\begin{figure}[H]
    \centering
    \begin{subfigure}[b]{0.49\textwidth}
        \includegraphics[width=\textwidth]{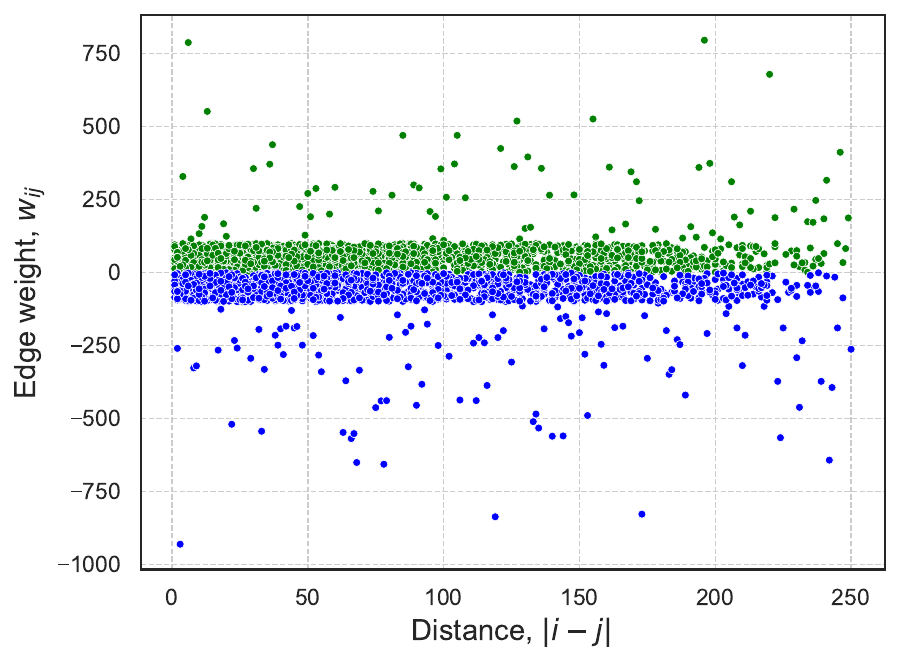}
        \subcaption{}
        \label{fig:newnewleft}
    \end{subfigure}
    \hfill
    \begin{subfigure}[b]{0.49\textwidth}
        \includegraphics[width=\textwidth]{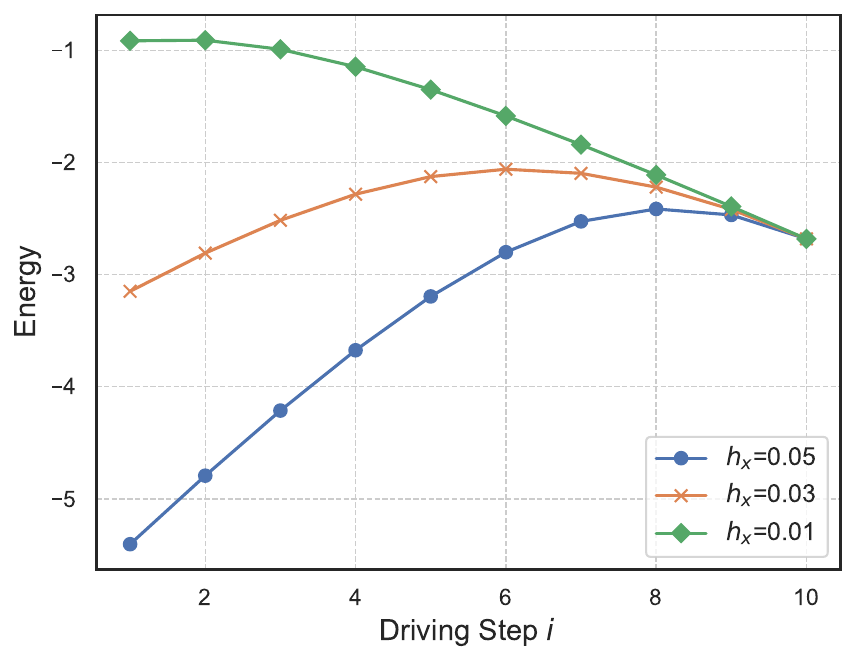}
        \subcaption{}
        \label{fig:newnewright}
    \end{subfigure}
    \caption{(\textbf{a}) The distribution of the edge weights $w_{i, j}$, sorted by $|i - j|$, the distance between the nodes. Most of the weights lie within the interval from $-100$ to $100$. (\textbf{b}) The variation of energy during the driving process for three different values of $h_x$. All three choices converge to the ground state, yielding the correct MaxCut value. The edge matrix has been normalized such that the maximum absolute value of its elements is $1$. Solving MaxCut instance bqp250-8 with $251$ nodes and $3265$ edges.}
    \label{fig:edgeweight}
\end{figure}

The third batch of instances includes even larger MaxCut problems with $N_V = 251$ and $N_E$ exceeding $3200$. The coupling between the nodes is so dense that these instances are not easily graphed or visualized. More importantly, in contrast to the examples discussed earlier, the edge weights in these instances vary over a wide range. For instance, the edge weight distribution of the instance bqp250-10 is shown in Figure~\ref{fig:edgeweight}. While there are a few outliers, the majority of $w_{i, j}$ fall within the range from $-100$ to $100$, regardless of $|i - j|$, the distance between nodes. Other instances listed in Table~\ref{tb:bqp} follow a similar distribution for $w_{i, j}$. As a result, the numerical scale of the QUBO matrix and the energy scale of the spin-spin interaction differ significantly from previous examples. To make the algorithm less sensitive to the overall numerical scale of $w_{i, j}$, we rescale the problem by dividing the matrix $w_{i, j}$ by the maximum absolute value of its elements. After this, $|w_{i, j}|$ becomes typically much smaller than $1$. Consequently, the value of $h_x$ is smaller compared to previous examples, as shown in the fifth column of Table~\ref{tb:bqp}.


\begin{table}[H]
\caption{The algorithm parameters used to successfully solve five MaxCut instances from the Biq Mac Library with 251 vertices and over 3200 edges, each with integer weights; $D = 30$, $M = 10$.}
\label{tb:bqp}
\begin{tabularx}{\textwidth}{
    >{\raggedright\arraybackslash}X||  
    >{\centering\arraybackslash}X|    
    >{\centering\arraybackslash}X|
    >{\centering\arraybackslash}X|
    >{\centering\arraybackslash}X|
    >{\centering\arraybackslash}X| 
    >{\centering\arraybackslash}X 
}
 \hline
Instance & $E_0$ & $N_V$ & $N_{E}$ & $h_x$ & $\eta$ & $N_{sw}$ \\ \hline
  bqp250-2	&-44810	&251		&3285 & 0.05 &0 & 10	\\
  bqp250-4 	&-41274	&251	 	&3397 & 0.05 &0 & 10	\\
  bqp250-6 	&-41014	&251		&3433 & 0.30 & 0 & 5		\\
  bqp250-8 	&-35726	&251		&3265 & 0.05 & 0 & 10	\\
  bqp250-10 	&-40442	&251		&3294 &1.00   & 0.3 & 5	\\
   \hline
\end{tabularx}
\end{table}

It's important to note that the choice of $h_x$ is not unique. Figure~\ref{fig:newnewright} compares the energy for three different values of $h_x$ during the driving process. All three configurations lead to the correct ground state, demonstrating the robustness of the algorithm, even with the significant increase in $N_E$ and $N_V$.


\subsection*{Summaries}

To summarize this section, we find it encouraging that the proposed quantum-inspired algorithm is capable of solving a wide variety of MaxCut problems of different sizes and types. However, success is not always guaranteed. Some cases prove to be more stubborn, resisting energy minimization. This is not unexpected, as similar challenges are frequently encountered in quantum annealing, where techniques like reverse and cyclic annealing have been introduced to prevent the system from becoming trapped in local minima. Refining and enhancing our algorithm to address such challenges will be part of future work. 

It is important to note that all QUBO problems can be formulated as MaxCut problems. Indeed, all the instances in Table~\ref{tb:bqp} were originally generated as QUBO problems by Beasley~\cite{beasly} and later translated into MaxCut format. Thus, the key takeaway from this section is that the quantum-inspired algorithm based on MPS and DMRG is versatile and continues to perform effectively on MaxCut (and, more broadly, QUBO) problems of larger sizes with denser connections.


\section{Discussions and Conclusions}
\label{sec:conclusions}


The emerging field of solving combinatorial optimization problems using quantum-inspired algorithms based on tensor networks is still in its early stages. Previous works have focused on the imaginary time evolution of MPS and experimented with PEPS~\cite{peps, portfolio}, where the main approach was to solve the problem Hamiltonian directly. Our algorithm, presented in Section~\ref{sec:algs}, differs in two significant ways. First, we introduce a driver/mixer Hamiltonian and work with a discrete driving schedule. Second, at each driving stage, we use DMRG to update the MPS wave function, progressively steering it toward the ground state of the problem Hamiltonian. In Sections~\ref{sec:experimentS} and~\ref{sec:experimentM}, we presented conclusive evidence that the algorithm can successfully solve a variety of Sudoku puzzles and MaxCut problems. To our knowledge, the combination of DMRG with discrete driving has not been demonstrated before in the context of QUBO problems, and the application of DMRG to Ising spin glass Hamiltonians with long-range interactions has not been systematically analyzed.

The algorithm proposed here is general and can be adapted to a wide range of QUBO problems. The problem sizes presented in Sections~\ref{sec:experimentS} and~\ref{sec:experimentM} (up to $251$ spins with $3,265$ couplings, which exceeds the capabilities of state-of-the-art QAOA) are kept relatively small, allowing most calculations to be completed on a laptop within a few minutes to an hour. The main goal of this work is to establish the feasibility of the proposed algorithm and test it across problems of diverse origin and size, while ensuring that the iterations remain fast and computationally inexpensive. In future work, we plan to explore how the algorithm can scale to tackle practical large-scale problems. For example, a popular choice for benchmarking many state-of-the-art MaxCut solvers is Gset, which includes problems with over 800 (up to 20,000) vertices~\cite{Gset}. For such large problems, different trials with varied initial MPS or driving parameters could be run in parallel on high-performance computing clusters over days or weeks.

Tensor network analysis of glassy spin Hamiltonians also provides new insights into algorithm design. Quantum annealing is inherently heuristic, and when faced with a new problem Hamiltonian, it is not immediately clear which driver Hamiltonian and annealing schedule will offer the best chance of reaching the global minimum. Without a precise understanding of the dynamics of interacting spins (qubits), the annealing process could result in either a success or failure. The many-spin wave functions obtained from tensor network calculations offer detailed information about the system under driving. This diagnostic data translates into a deeper understanding of the role of each parameter, aiding in more informed decisions when selecting algorithmic parameters. The ability to directly access the quantum states is a significant strength of the quantum-inspired approach.

Quantum optimization algorithms, such as quantum annealing and QAOA, hold great potential to surpass classical algorithms for industrial-scale optimization problems. In the near term, however, their performance will be limited by the capabilities of current hardware. Before these quantum algorithms can reach their full potential, quantum-inspired algorithms can bridge the gap by solving middle- to large-scale problems within polynomial time via quantum simulations on classical computers. A few recent examples, some of which are not directly related to optimization tasks, illustrate the potential of tensor network calculations. IBM's $127$-qubit Eagle processor initially showed experimental results on two-dimensional quantum spin systems that were believed to have achieved quantum advantage -- i.e., performing ``accurate computations at a scale beyond brute-force classical simulation''~\cite{Kim:2023aa}. However, soon after, improved tensor network calculations on classical computers produced more accurate results than the quantum processor~\cite{PRXQuantum.5.010308}. Tensor networks have also been used to replace parts of the QAOA~\cite{streif2020} or quantum annealing~\cite{luchnikov2024} algorithms. In yet another recent development, several quantum-inspired solvers based on tensor trains were demonstrated to successfully solve nonlinear partial differential equations~\cite{tt1, tt2, tt3, tt4}. These examples further underscore that tensor network methods are powerful and versatile tools, extending beyond combinatorial optimization problems. The near-term landscape of optimization will likely feature a coexistence where quantum-inspired algorithms and pure quantum algorithms compete and complement each other.

\appendix

\section[\appendixname~\thesection]{Matrix Product States and Density Matrix Renormalization Group}
\label{appendix:A}


In recent years, tensor networks have emerged as a powerful tool not only in quantum many-body physics and quantum chemistry but also in quantum information and machine learning. One of the most well-understood tensor networks is Matrix Product States (MPS), also known as tensor trains~\cite{verstraete,tensorreview}. In our context, the quantum state of an $N$-spin system can be expanded as
\begin{equation}
\ket{\psi} = \sum_{{s_i}} T^{s_1, s_2, \ldots, s_N} \ket{s_1, s_2, \ldots, s_N},
\end{equation}
where $\ket{s_1, s_2, \ldots, s_N}$ represents the product state in which the $i$th spin is in the eigenstate $\ket{s_i}$ with eigenvalue $s_i = \pm \frac{1}{2}$, and $T^{s_1, s_2, \ldots, s_N}$ is a rank-$N$ tensor with $N$ free indices. (In this appendix, we use $N$ instead of $N_s$ to denote the number of spins, simplifying the index notation and avoiding clutter.) For large $N$, directly manipulating $T$ becomes prohibitively expensive, as the number of its components, $2^N$, grows exponentially. The key idea behind MPS is to express $T$ as the product of a sequence of lower-rank tensors:
\begin{equation}
T^{s_1, s_2, \ldots, s_N} = \sum_{{\alpha_i}} A^{s_1}_{\alpha_1} A^{s_2}_{\alpha_1, \alpha_2} A^{s_3}_{\alpha_2, \alpha_3} \ldots A^{s_N}_{\alpha_{N - 1}}
\label{mps}
\end{equation}

Each object $A$ has one upper index $s_i$. For a given $s_i$, $A^{s_i}$ can be viewed as a matrix, since it carries two lower indices, which serve as row and column labels. (Note that $A^{s_1}$ and $A^{s_N}$ are special, as they only have one lower index and correspond to row and column matrices, respectively.) These lower indices are summed over in equation~\eqref{mps} within the range $\alpha_i = 1, 2, \ldots, D$, where $D$ is the bond dimension, a finite number. To illustrate how multiple-spin quantum states can be constructed from MPS, consider the simple example with $N = 4$ and $D = 2$~\cite{PhysRevA.78.012356}:
\begin{align*}
\ket{\psi} & =
\begin{pmatrix}
\ket{\uparrow} &
\ket{\downarrow}
\end{pmatrix}
\begin{pmatrix}
\ket{\uparrow} & \ket{\downarrow}\\
0 & \ket{\uparrow}
\end{pmatrix}
\begin{pmatrix}
\ket{\uparrow} & \ket{\downarrow}\\
0 & \ket{\uparrow}
\end{pmatrix}
\begin{pmatrix}
\ket{\downarrow}\\
\ket{\uparrow}
\end{pmatrix}
= \ket{\uparrow \uparrow \uparrow \downarrow} + \ket{\uparrow \uparrow \downarrow \uparrow} + \ket{\uparrow \downarrow \uparrow \uparrow} + \ket{\downarrow \uparrow \uparrow \uparrow}
\end{align*}

The four-spin state $\ket{\psi}$ is obtained from the product of four matrices, hence the name ``matrix product states.'' The MPS representation in equation~\eqref{mps} is exact, provided that $D$ is sufficiently large. In practice, a finite value of $D$ can be chosen so that MPS serves as a good approximation to $T$. A major advantage of using MPS is that the total number of components scales as 
$N D^2$, representing a significant reduction compared to the $2^N$ components required for the full tensor. 

In certain cases, theoretical lower bounds on the required bond dimensions are known. For one-dimensional (1D) gapped Hamiltonians with local interactions, the ground state can be efficiently approximated by a MPS with sublinear bond dimension $$D = \exp(\tilde{O}\left(\log^{3/4} N / \Delta^{1/4}\right)),$$ where $\Delta$ denotes the spectral gap and $N$ is the system size~\cite{hastings2007area, arad2013area}. Moreover, a polynomial-time algorithm exists to find a good MPS approximation of the ground state~\cite{PhysRevA.76.030307}. For 1D gapless (critical) systems, MPS with polynomial bond dimension can still approximate the ground state accurately, provided that certain Rényi entropies grow at most logarithmically with system size~\cite{PhysRevB.73.094423, PhysRevLett.100.030504}. Less is known about systems with long-range interactions, but the area law appears to hold for specific cases with power-law decaying interactions~\cite{kuwahara2020area}.

The ground states of many one-dimensional quantum spin systems can be found accurately using the MPS representation through an algorithm called Density Matrix Renormalization Group (DMRG)~\cite{white, dmrgreview}. In its modern form, DMRG can be viewed as an iterative variational procedure that optimizes the MPS tensors to minimize the energy expectation value. Roughly speaking, the algorithm starts with an initial set of $A^{s_i}$ tensors and focuses on one bond at a time, i.e., two neighboring matrices 
$A^{s_i}$ and $A^{s_{i + 1}}$. The components of these two matrices are adjusted by applying the Hamiltonian, treating the rest of the system as an effective environment, and solving a linear algebra problem; for more details, see~\cite{dmrgreview}. Once these two tensors are updated, the algorithm moves on to the next bond, sweeping from left to right and then back. The number of sweeps required to converge to the ground state depends on the Hamiltonian, the choice of the initial MPS, and DMRG parameters such as $D$ and other fine-tuning settings for the iterative eigensolver. The computational cost of the DMRG algorithm typically scales as $N D^3$. We choose DMRG as the primary method for solving QUBO problems because it is well-established and available in several open-source libraries. Additionally, we find that it outperforms other MPS-based algorithms in terms of speed and stability.

We briefly comment on how the bond dimension $D$ is chosen in practice. It is helpful to view DMRG algorithm through the lens of variational optimization. Recall that the quality of the variational many-body wavefunction, represented as a MPS, depends crucially on the bond dimension $D$. The success of the variational procedure also hinges on the details of the update strategy, including the choice of the initial MPS, the number of sweeps (\texttt{nsweeps}), and other algorithmic parameters such as truncation thresholds, the number of iterations in the linear solver, and the use of noise~\cite{itensor}. A careful DMRG simulation typically follows a so-called ``parameter schedule,'' in which $D$, \texttt{nsweeps}, and related parameters are adjusted dynamically. For example, the bond dimension is often increased gradually over successive sweeps to improve convergence and accuracy~\cite{itensor}.

Quantum-inspired approaches to QUBO problems based on tensor networks were recently explored by Patra et al.~\cite{peps} and Mugel et al.~\cite{portfolio}. These authors seek to find the ground states of Ising spin glass models, analogous to~\eqref{SGH} here, by describing many-spin states using either MPS or its two-dimensional generalization, Projected Entangled Pair States (PEPS)~\cite{verstraete}. Starting from a guess state $\ket{\psi_0}$, imaginary-time evolution is applied iteratively in small Trotter steps $\delta \tau$, $\ket{\psi_{i + 1}} = e^{-H \delta \tau} \ket{\psi_i}$, to approach the ground state. A drawback of this approach is its significant computational cost. During the time evolution of MPS, if the Hamiltonian $H$ contains dense, long-range couplings such as $J_{m, n}$, at each Trotter step applying each operator $e^{-\delta \tau J_{m, n} S_m S_n}$ involves a sequence of swap operations that grow with $d = |m - n|$, which significantly slows down the calculation. Updating PEPS for glassy and densely connected Hamiltonians presents a formidable technical challenge. It is currently unclear whether the recently proposed flexible PEPS updating scheme~\cite{peps} can efficiently reach the global minimum for hundreds of spins. For these reasons, in contrast to~\cite{peps,portfolio}, we avoid imaginary time evolution and PEPS.

It is important to emphasize once again that our strategy is \emph{not} to directly target the ground state of the Ising spin glass Hamiltonian using DMRG with a large bond dimension $D$ and a standard parameter schedule. As discussed in the main text, such an approach rarely succeeds, as DMRG often gets trapped in local minima. Instead, our scheme focuses on constructing a high-quality initial MPS -- a highly entangled superposition of spin configurations -- by following a discrete driving schedule. At each step, DMRG brings the MPS close to the ground state of a sequence of intermediate (mixed) Hamiltonians. Once this initial MPS (typically with $D = 60$) is obtained, the final DMRG optimization for the target Ising spin glass Hamiltonian becomes significantly more efficient. In fact, the bond dimension typically decreases with each sweep, eventually converging to a product state with $D = 1$. This behavior contrasts sharply with conventional DMRG workflows, where $D$ is often increased during the calculation. We hope our approach will inspire further creative applications of DMRG to a broader class of complex optimization problems.


\section{An Analogy for MPS-DMRG-Based Search}
\label{appendix:B}


For readers unfamiliar with the inner workings of MPS and DMRG, the core idea of our algorithm can be grasped through a rough analogy. Our task is to search for the global minimum in a rugged landscape. To accomplish this, we employ a group of ``sweepers'' -- experts in exploring local terrains and finding local energy minima -- and charter a helicopter to airlift them from one location to the next.

The search begins at a chosen starting point $\mathbf{r}_1 = (a_1, b_1)$, where the local topography is either known or easy to explore (e.g., the ground state of $H_1 = H_x$ is already known). Equipped with better insights (provided by the MPS wave function) from the local sweeps conducted in the previous step, the team then ``hops'' to a new location $\mathbf{r}_2 = (a_2, b_2)$, some distance away, via helicopter. The new terrain may be more rugged, but the sweepers now have an educated guess of the starting point, improving their chances of reaching the new local minimum.

The iterative search strategy can be summarized as follows:
\begin{algorithm}
\begin{algorithmic}
\REPEAT

\STATE hop, sweep

\UNTIL{$M$ times}
\end{algorithmic}
\end{algorithm}

The helicopter makes $M$ stops, and at each location $\mathbf{r}_i$, the sweepers explore a different $H_i$ and update $\ket{\psi_i}$. As we demonstrate in Section~\ref{sec:experimentS} and Section~\ref{sec:experimentM}, the algorithm performs surprisingly well, often reaching the true ground state within just $M = 5$ hops, with $5$ DMRG sweeps at each location.

This analogy, while useful, is somewhat misleading. It is important to remember that the DMRG sweeps at each stop are quantum mechanical operations aimed at refining the MPS wave functions. The term ``terrain'' in the analogy refers to the high-dimensional variational parameter space of the MPS wave functions, with a dimension on the order of $\sim N_s D^2$. In reality, the problem is much more complex than exploring three-dimensional terrains. Fortunately, we have an efficient and well-established tool in DMRG to handle this task.

Another key ingredient of our algorithm is the inclusion of the driver term $H_x$ in each $H_i$. Recall that $H_x$ does not commute with $H_z$, and without it, the system would be purely classical. This term facilitates spin flips and quantum tunneling, and it enables quantum parallelism. Each MPS represents a superposition of many spin states, which are evolved simultaneously. Thanks to DMRG's efficiency in optimizing the MPS, the driving schedule can afford to be discrete, allowing the algorithm to reach the global minimum of $H_z$ in just a few steps. This stands in contrast to traditional quantum annealing, where qubits are carefully controlled to evolve adiabatically along a continuous path.




\begin{thebibliography}{99}

\bibitem[1]{papadimitriou1998combinatorial}
Christos~H Papadimitriou and Kenneth Steiglitz.
\newblock \emph{Combinatorial optimization: algorithms and complexity}.
\newblock Dover Publications, 1998 

\bibitem[2]{korte2011combinatorial}
Bernhard~H Korte, Jens Vygen, B~Korte, and J~Vygen.
\newblock \emph{Combinatorial optimization}, volume~1.
\newblock Springer, 2011.

\bibitem[3]{cormen2022introduction}
Thomas~H Cormen, Charles~E Leiserson, Ronald~L Rivest, and Clifford Stein.
\newblock \emph{Introduction to algorithms}.
\newblock MIT press, 2022.

\bibitem[4]{Glover:2022aa}
Fred Glover, Gary Kochenberger, Rick Hennig, and Yu~Du.
\newblock {Quantum bridge analytics I: a tutorial on formulating and using
  {QUBO} models}.
\newblock \emph{Annals of Operations Research}, 314(1):141--183, 2022.

\bibitem[5]{punnen2022quadratic}
Abraham~P Punnen, editor.
\newblock \emph{The quadratic unconstrained binary optimization problem}.
\newblock Springer Cham, 2022.

\bibitem[6]{Barahona_1982}
Francisco Barahona. 
\newblock On the computational complexity of {I}sing spin glass models.
\newblock \emph{Journal of Physics A: Mathematical and General}, 15(10):3241,
  oct 1982.

\bibitem[7]{Abbas:2024aa}
Amira Abbas, Andris Ambainis, Brandon Augustino, Andreas B{\"a}rtschi, Harry
  Buhrman, Carleton Coffrin, Giorgio Cortiana, Vedran Dunjko, Daniel~J. Egger,
  Bruce~G. Elmegreen, Nicola Franco, Filippo Fratini, Bryce Fuller, Julien
  Gacon, Constantin Gonciulea, Sander Gribling, Swati Gupta, Stuart Hadfield,
  Raoul Heese, Gerhard Kircher, Thomas Kleinert, Thorsten Koch, Georgios
  Korpas, Steve Lenk, Jakub Marecek, Vanio Markov, Guglielmo Mazzola, Stefano
  Mensa, Naeimeh Mohseni, Giacomo Nannicini, Corey O'Meara, Elena~Pe{\~n}a
  Tapia, Sebastian Pokutta, Manuel Proissl, Patrick Rebentrost, Emre Sahin,
  Benjamin C.~B. Symons, Sabine Tornow, V{\'\i}ctor Valls, Stefan Woerner,
  Mira~L. Wolf-Bauwens, Jon Yard, Sheir Yarkoni, Dirk Zechiel, Sergiy Zhuk, and
  Christa Zoufal.
\newblock Challenges and opportunities in quantum optimization.
\newblock \emph{Nature Reviews Physics}, 6(12):718--735, 2024.

\bibitem[8]{10.3389/fphy.2014.00005}
Andrew Lucas.
\newblock Ising formulations of many {NP} problems.
\newblock \emph{Frontiers in Physics}, 2, 2014.

\bibitem[9]{Mohseni:2022aa}
Naeimeh Mohseni, Peter~L. McMahon, and Tim Byrnes.
\newblock Ising machines as hardware solvers of combinatorial optimization
  problems.
\newblock \emph{Nature Reviews Physics}, 4(6):363--379, 2022.

\bibitem[10]{RevModPhys.80.1061}
Arnab Das and Bikas~K. Chakrabarti.
\newblock Colloquium: Quantum annealing and analog quantum computation.
\newblock \emph{Rev. Mod. Phys.}, 80:1061--1081, Sep 2008.

\bibitem[11]{finnila1994quantum}
Aleta~Berk Finnila, Maria~A Gomez, C~Sebenik, Catherine Stenson, and Jimmie~D
  Doll.
\newblock Quantum annealing: A new method for minimizing multidimensional
  functions.
\newblock \emph{Chemical Physics Letters}, 219(5-6):343--348, 1994.

\bibitem[12]{PhysRevE.58.5355}
Tadashi Kadowaki and Hidetoshi Nishimori.
\newblock Quantum annealing in the transverse {I}sing model.
\newblock \emph{Phys. Rev. E}, 58:5355--5363, Nov 1998.

\bibitem[13]{farhi2014quantum}
Edward Farhi, Jeffrey Goldstone, and Sam Gutmann.
\newblock A quantum approximate optimization algorithm.
\newblock \emph{arXiv preprint arXiv:1411.4028}, 2014.

\bibitem[14]{hadfield2019quantum}
Stuart Hadfield, Zhihui Wang, Bryan O'gorman, Eleanor~G Rieffel, Davide
  Venturelli, and Rupak Biswas.
\newblock From the quantum approximate optimization algorithm to a quantum
  alternating operator ansatz.
\newblock \emph{Algorithms}, 12(2):34, 2019.

\bibitem[15]{advantage}
The {D}-wave {A}dvantage system: An overview.
\newblock
Available at:  \url{https://www.dwavequantum.com/media/s3qbjp3s/14-1049a-a_the_d-wave_advantage_system_an_overview.pdf} (Accessed on date: 1 September 2024).

\bibitem[16]{blekos2024review}
Kostas Blekos, Dean Brand, Andrea Ceschini, Chiao-Hui Chou, Rui-Hao Li, Komal
  Pandya, and Alessandro Summer.
\newblock A review on quantum approximate optimization algorithm and its
  variants.
\newblock \emph{Physics Reports}, 1068:1--66, 2024.

\bibitem[17]{pagano2020quantum}
Guido Pagano, Aniruddha Bapat, Patrick Becker, Katherine~S Collins, Arinjoy De,
  Paul~W Hess, Harvey~B Kaplan, Antonis Kyprianidis, Wen~Lin Tan, Christopher
  Baldwin, et~al.
\newblock Quantum approximate optimization of the long-range {I}sing model with
  a trapped-ion quantum simulator.
\newblock \emph{Proceedings of the National Academy of Sciences},
  117(41):25396--25401, 2020.

\bibitem[18]{du2025new}
Yu~Du, Haibo Wang, Rick Hennig, Amit Hulandageri, Gary Kochenberger, and Fred
  Glover.
\newblock New advances for quantum-inspired optimization.
\newblock \emph{International Transactions in Operational Research},
  32(1):6--17, 2025.

\bibitem[19]{pittmine}
Tianyi Hao, Xuxin Huang, Chunjing Jia, and Cheng Peng.
\newblock A quantum-inspired tensor network algorithm for constrained
  combinatorial optimization problems.
\newblock \emph{Frontiers in Physics}, 10, 2022.

\bibitem[20]{inspiredreview}
Larry Huynh, Jin Hong, Ajmal Mian, Hajime Suzuki, Yanqiu Wu, and Seyit Camtepe.
\newblock Quantum-inspired machine learning: a survey.
\newblock \emph{arXiv preprint arXiv:2308.11269}, 2023.

\bibitem[21]{portfolio}
Samuel Mugel, Carlos Kuchkovsky, Escol\'astico S\'anchez, Samuel
  Fern\'andez-Lorenzo, Jorge Luis-Hita, Enrique Lizaso, and Rom\'an Or\'us.
\newblock Dynamic portfolio optimization with real datasets using quantum
  processors and quantum-inspired tensor networks.
\newblock \emph{Phys. Rev. Res.}, 4:013006, Jan 2022.

\bibitem[22]{mucke}
Sascha M\"{u}cke.
\newblock A simple {QUBO} formulation of sudoku.
\newblock In \emph{Proceedings of the Genetic and Evolutionary Computation
  Conference Companion}, GECCO '24 Companion, pages 1958--1962, New York, NY,
  USA, 2024. Association for Computing Machinery.

\bibitem[23]{verstraete}
Frank Verstraete, Valentin Murg, and Juan Ignacio Cirac. 
\newblock Matrix product states, projected entangled pair states, and
  variational renormalization group methods for quantum spin systems.
\newblock \emph{Advances in Physics}, 57(2):143--224, 2008.

\bibitem[24]{dmrgreview}
Ulrich Schollw\"ock. 
\newblock The density-matrix renormalization group.
\newblock \emph{Rev. Mod. Phys.}, 77:259--315, Apr 2005.

\bibitem[25]{white}
Steven~R. White.
\newblock Density matrix formulation for quantum renormalization groups.
\newblock \emph{Phys. Rev. Lett.}, 69:2863--2866, Nov 1992.

\bibitem[26]{tensorreview}
Rom{\'a}n Or{\'u}s.
\newblock A practical introduction to tensor networks: Matrix product states
  and projected entangled pair states.
\newblock \emph{Annals of Physics}, 349:117--158, 2014.

\bibitem[27]{yato2003complexity}
Takayuki Yato and Takahiro Seta.
\newblock Complexity and completeness of finding another solution and its
  application to puzzles.
\newblock \emph{IEICE Transactions on Fundamentals of Electronics, Communications and Computer Sciences}, 86(5):1052--1060, 2003.

\bibitem[28]{UedaN96}
Nobuhisa Ueda and Tadaaki Nagao. 
\newblock {NP}-completeness results for {NONOGRAM} via parsimonious reductions.
\newblock \emph{Technical Report TR96-0008, Department of Computer Science,
  Tokyo Institute of Technology}, 1996.

\bibitem[29]{PhysRevLett.35.1792}
David Sherrington and Scott Kirkpatrick.
\newblock Solvable model of a spin-glass.
\newblock \emph{Phys. Rev. Lett.}, 35:1792--1796, Dec 1975.

\bibitem[30]{Van:1977aa}
J~Vannimenus and G~Toulouse.
\newblock Theory of the frustration effect. {II}. {I}sing spins on a square
  lattice.
\newblock \emph{Journal of Physics C: Solid State Physics}, 10(18):L537, 1977.

\bibitem[31]{peps}
Siddhartha Patra, Sukhbinder Singh, and Rom\'an Or\'us.
\newblock Projected entangled pair states with flexible geometry.
\newblock \emph{Phys. Rev. Res.}, 7:L012002, Jan 2025.

\bibitem[32]{itensor}
Matthew Fishman, Steven~R. White, and E.~Miles Stoudenmire.
\newblock {The ITensor Software Library for Tensor Network Calculations}.
\newblock \emph{{SciPost Phys. Codebases}}, 4, 2022.

\bibitem[33]{pimc2002}
Roman Marto\ifmmode~\check{n}\else \v{n}\fi{}\'ak, Giuseppe~E. Santoro, and
  Erio Tosatti.
\newblock Quantum annealing by the path-integral {Monte Carlo} method: The
  two-dimensional random {Ising} model.
\newblock \emph{Phys. Rev. B}, 66:094203, 2002.

\bibitem[34]{Santoro2002}
Giuseppe E. Santoro, Roman Martonak, Erio Tosatti, and Roberto Car. 
\newblock Theory of quantum annealing of an {Ising} spin glass.
\newblock \emph{Science}, 295(5564):2427--2430, 2002.

\bibitem[35]{nycom}
Available at: \url{https://www.nytimes.com/puzzles/sudoku} (Accessed on date: 1 September 2024 
).

\bibitem[36]{nynet}
Available at: \url{https://nytsudoku.net} (Accessed on date: 1 September 2024).

\bibitem[37]{Karp1972}
Richard~M. Karp.
\newblock \emph{Reducibility among Combinatorial Problems}, pages 85--103.
\newblock Springer US, Boston, MA, 1972.

\bibitem[38]{spglassmaxcut}
Francisco Barahona, Martin Gr\"{o}tschel, Michael J\"{u}nger, and Gerhard
  Reinelt.
\newblock An application of combinatorial optimization to statistical physics
  and circuit layout design.
\newblock \emph{Operations Research}, 36(3):493--513, 1988.

\bibitem[39]{biqmac}
Available at: \url{https://biqmac.aau.at/} (Accessed on date: 1 September 2024).

\bibitem[40]{instances}
Available at:\url{http://bqp.cs.uni-bonn.de/library/html/instances.html} (Accessed on date: 1 September 2024).

\bibitem[41]{beasly}
Available at: \url{https://people.brunel.ac.uk/~mastjjb/jeb/info.html} (Accessed on date: 1 September 2024).

\bibitem[42]{Gset}
Available at: \url{https://web.stanford.edu/~yyye/yyye/Gset/} (Accessed on date: 1 September 2024).

\bibitem[43]{Kim:2023aa}
Youngseok Kim, Andrew Eddins, Sajant Anand, Ken~Xuan Wei, Ewout van~den Berg,
  Sami Rosenblatt, Hasan Nayfeh, Yantao Wu, Michael Zaletel, Kristan Temme, and
  Abhinav Kandala.
\newblock Evidence for the utility of quantum computing before fault tolerance.
\newblock \emph{Nature}, 618(7965):500--505, 2023.

\bibitem[44]{PRXQuantum.5.010308}
Joseph Tindall, Matthew Fishman, E.~Miles Stoudenmire, and Dries Sels.
\newblock Efficient tensor network simulation of {IBM's} {Eagle} kicked {Ising}
  experiment.
\newblock \emph{PRX Quantum}, 5:010308, Jan 2024.

\bibitem[45]{streif2020}
Michael Streif and Martin Leib.
\newblock Training the quantum approximate optimization algorithm without
  access to a quantum processing unit.
\newblock \emph{Quantum Science and Technology}, 5(3):034008, 2020.

\bibitem[46]{luchnikov2024}
Ilia~A. Luchnikov, Egor~S. Tiunov, Tobias Haug, and Leandro Aolita.
\newblock Large-scale quantum annealing simulation with tensor networks and
  belief propagation.
\newblock \emph{arXiv preprint arXiv:2409.12240}, 2024.

\bibitem[47]{tt2}
Aleix Bou-Comas, Marcin Płodzień, Luca Tagliacozzo, and Juan~José
  García-Ripoll.
\newblock {Quantics Tensor Train for solving {Gross-Pitaevskii} equation}.
\newblock \emph{arXiv preprint arXiv:2507.03134}, 2025.

\bibitem[48]{tt4}
Qian-Can Chen, I-Kang Liu, Jheng-Wei Li, and Chia-Min Chung.
\newblock {Solving the {Gross-Pitaevskii} Equation with Quantic Tensor Trains:
  Ground States and Nonlinear Dynamics}.
\newblock \emph{arXiv preprint arXiv:2507.04279}, 2025.

\bibitem[49]{tt1}
Ryan J.~J. Connor, Callum~W. Duncan, and Andrew~J. Daley.
\newblock {Tensor network methods for the {Gross-Pitaevskii} equation on fine
  grids}.
\newblock \emph{arXiv preprint arXiv:2507.01149}, 2025.

\bibitem[50]{tt3}
Marcel Niedermeier, Adrien Moulinas, Thibaud Louvet, Jose~L. Lado, and Xavier
  Waintal.
\newblock {Solving the {Gross-Pitaevskii} equation on multiple different scales
  using the quantics tensor train representation}.
\newblock \emph{arXiv preprint arXiv:2507.04262}, 2025.

\bibitem[51]{PhysRevA.78.012356}
Gregory~M. Crosswhite and Dave Bacon.
\newblock Finite automata for caching in matrix product algorithms.
\newblock \emph{Phys. Rev. A}, 78:012356, Jul 2008.


\bibitem[52]{arad2013area}
Itai Arad, Alexei Kitaev, Zeph Landau, and Umesh Vazirani.
\newblock An area law and sub-exponential algorithm for {1D} systems.
\newblock \emph{arXiv preprint arXiv:1301.1162}, 2013.

\bibitem[53]{hastings2007area}
Matthew~B Hastings.
\newblock An area law for one-dimensional quantum systems.
\newblock \emph{Journal of statistical mechanics: theory and experiment},
  2007(08):P08024, 2007.

\bibitem[54]{PhysRevA.76.030307}
Alastair Kay.
\newblock {Quantum-Merlin-Arthur-complete translationally invariant Hamiltonian
  problem and the complexity of finding ground-state energies in physical
  systems}.
\newblock \emph{Phys. Rev. A}, 76:030307, Sep 2007.

\bibitem[55]{PhysRevLett.100.030504}
Norbert Schuch, Michael~M. Wolf, Frank Verstraete, and Juan Ignacio Cirac. 
\newblock Entropy scaling and simulability by matrix product states.
\newblock \emph{Phys. Rev. Lett.}, 100:030504, Jan 2008.

\bibitem[56]{PhysRevB.73.094423}
Frank Verstraete and Juan Ignacio Cirac. 
\newblock Matrix product states represent ground states faithfully.
\newblock \emph{Phys. Rev. B}, 73:094423, Mar 2006.

\bibitem[57]{kuwahara2020area}
Tomotaka Kuwahara and Keiji Saito.
\newblock Area law of noncritical ground states in {1D} long-range interacting
  systems.
\newblock \emph{Nature Communications}, 11(1):4478, 2020.

\end{thebibliography}


\end{document}